%% file: main.tex
\documentclass[10pt,twocolumn,letterpaper]{article}

\usepackage{cvpr}              
\input{preamble}
\definecolor{cvprblue}{rgb}{0.21,0.49,0.74}
\usepackage[pagebackref,breaklinks,colorlinks,allcolors=cvprblue]{hyperref}

\usepackage{hyperref}
\usepackage{url}

\usepackage[utf8]{inputenc} 
\usepackage[T1]{fontenc}    
\usepackage{hyperref}       
\usepackage{url}            
\usepackage{booktabs}       
\usepackage{amsfonts}       
\usepackage{nicefrac}       
\usepackage{microtype}      
\usepackage{xcolor}         

\usepackage{colortbl}
\usepackage{xcolor}
\definecolor{lightgray}{gray}{0.9}
\usepackage{booktabs}
\usepackage{color, soul}
\usepackage{graphicx}
\usepackage{amsmath}
\usepackage{empheq}
\usepackage{tcolorbox}
\usepackage{moresize}
\usepackage{caption}
\usepackage{subcaption} 

\usepackage{amssymb}  
\usepackage{pifont}   

\usepackage{algorithm}
\usepackage{algpseudocode}

\usepackage{listings}
\usepackage{pythonhighlight}

\usepackage{makecell}
\usepackage{fontawesome5}
\newcommand{\trained}{\textcolor{red!70}{\faFire}}         
\newcommand{\frozen}{\textcolor{blue!70}{\faSnowflake}}    

\usepackage{siunitx}    

\usepackage{array}      
\usepackage{multirow}   
\usepackage{graphicx}   
\usepackage{caption}    
\usepackage{multicol}   
\usepackage{threeparttable} 

\usepackage{tabularx}

\newcommand\freefootnote[1]{%
  \let\svthefootnote\thefootnote
  \let\thefootnote\relax
  \footnotetext{\textsuperscript{*}#1}
  \let\thefootnote\svthefootnote
}

\definecolor{lightgray}{gray}{0.9}
\definecolor{lightblue}{rgb}{0.93,0.95,1.0}
\definecolor{darkgreen}{rgb}{0.0,0.6,0.0}
\definecolor{darkblue}{rgb}{0.0,0.0,0.5}
\definecolor{pinegreen}{rgb}{0.0, 0.47, 0.44}
\definecolor{deepmagenta}{rgb}{0.8, 0.0, 0.8}
\definecolor{amber}{rgb}{1.0, 0.49, 0.0}
\definecolor{Gray}{gray}{0.9}

\usepackage{soul} 

\definecolor{uscgold}{HTML}{d9ae02}      
\definecolor{carnegiered}{HTML}{C41230}  
\definecolor{berkeleyblue}{HTML}{003057} 

\newcommand{\ignorebig}[1]{}

\def\Secref#1{Section~\ref{#1}}

\newcommand{\minisection}[1]{\noindent{\textbf{#1}.}}

\newlength\savewidth

\definecolor{citecolor}{RGB}{34,139,34}
\definecolor{lightred}{RGB}{241,140,142}
\definecolor{amber(sae/ece)}{rgb}{1.0, 0.49, 0.0}
\definecolor{battleshipgrey}{rgb}{0.52, 0.52, 0.51}
\definecolor{cadmiumorange}{rgb}{0.93, 0.53, 0.18}
\definecolor{applegreen}{rgb}{0.55, 0.71, 0.0}
\definecolor{cadmiumgreen}{rgb}{0.0, 0.42, 0.24}
\definecolor{forestgreen}{rgb}{0.13, 0.55, 0.13}
\definecolor{red}{rgb}{0.89, 0.0, 0.13}

\def\paperID{17666} 
\def\confName{CVPR}
\def\confYear{2026}

\title{\raisebox{-.25\height}{\includegraphics[height=1.6cm]{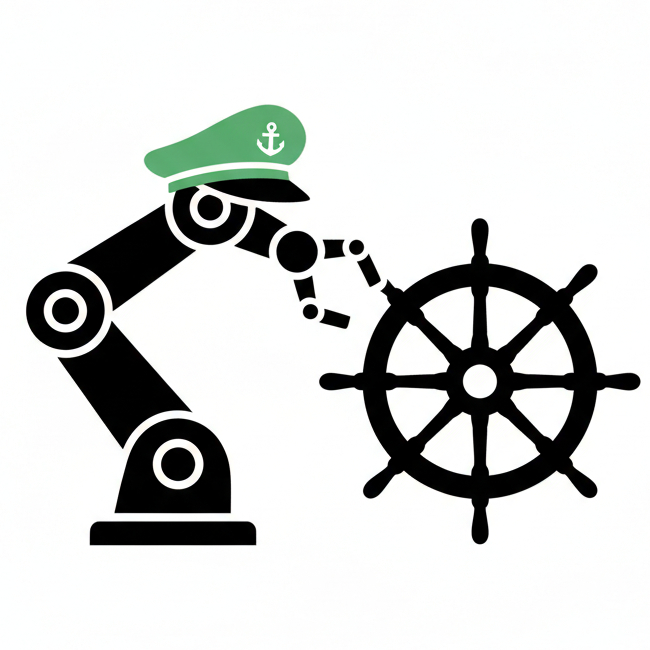}}Mechanistic Finetuning of Vision-Language-Action Models \\ via Few-Shot Demonstrations}

\author{
    Chancharik Mitra\textsuperscript{1*} \quad 
    Yusen Luo\textsuperscript{2*} \quad 
    Raj Saravanan\textsuperscript{3*} \quad 
    Dantong Niu\textsuperscript{3} \quad 
    Anirudh Pai\textsuperscript{3} \\
    Jesse Thomason\textsuperscript{2} \quad 
    Trevor Darrell\textsuperscript{3} \quad 
    Abrar Anwar\textsuperscript{2} \quad 
    Deva Ramanan\textsuperscript{1} \quad 
    Roei Herzig\textsuperscript{3, 4} \\ \\
{\small\color{black!70} \textsuperscript{1}Carnegie Mellon University \quad 
\textsuperscript{2}University of Southern California \quad
\textsuperscript{3}University of California, Berkeley \quad 
\textsuperscript{4}MIT-IBM Watson AI Lab}
}
\begin{document}
\maketitle
\input{sec/0_abstract}
\input{fig_1_teaser}
\input{sec/1_introduction_v3}
\input{fig_2_main}
\input{sec/2_related_works}
\input{tbl_1_main}

\input{sec/3_methods}

\input{sec/4_setup_details}

\input{sec/5_experiments_and_results}

\input{sec/6_conclusion_and_limitations}

{
    \small
    \bibliographystyle{ieeenat_fullname}
    \bibliography{main}
}

\input{sec/7_supp}

\end{document}

%% file: sec/0_abstract.tex
\begin{abstract}
Vision-Language Action (VLAs) models promise to extend the remarkable success of vision-language models (VLMs) to robotics. 
Yet, unlike VLMs in the vision-language domain, VLAs for robotics require finetuning to contend with varying physical factors like robot embodiment, environment characteristics, and spatial relationships of each task.
Existing fine-tuning methods lack specificity, adapting the same set of parameters regardless of a task's visual, linguistic, and physical characteristics.
Inspired by functional specificity in neuroscience, we hypothesize that it is more effective to finetune sparse model representations specific to a given task.  
In this work, we introduce \textbf{Robotic Steering}, a finetuning approach grounded in mechanistic interpretability that leverages few-shot demonstrations to identify and selectively finetune task-specific attention heads aligned with the physical, visual, and linguistic requirements of robotic tasks.
Through comprehensive on-robot evaluations with a Franka Emika robot arm, we demonstrate that \textbf{Robotic Steering} outperforms LoRA while achieving superior robustness under task variation, reduced computational cost, and enhanced interpretability for adapting VLAs to diverse robotic tasks. Project Page: \url{https://chancharikmitra.github.io/robosteering/}

\end{abstract}

%% file: fig_1_teaser.tex
\begin{figure}[t]
    \centering
    \includegraphics[width=\linewidth]{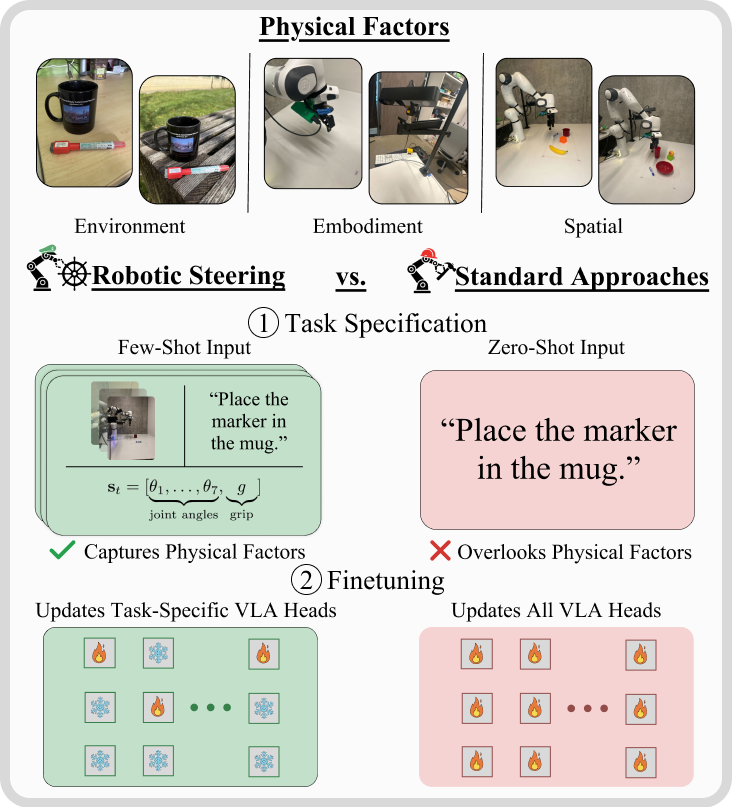}
    \caption{\textbf{Robotic Steering.} (left) leverages few-shot examples that capture the inherent physical variability (top) of robotic tasks to identify and selectively finetune only task-relevant attention heads. In contrast, standard approaches (right) specify tasks only with a language expression and update all parameters. Because of this, our novel method is a more performant, efficient, and generalizable approach for few-shot VLA finetuning.}
    \label{fig:teaser}
\end{figure}


%% file: sec/1_introduction_v3.tex
\section{Introduction}


\begin{quote}
\raggedright 
{\em "It is tempting, if the only tool you have is a hammer, to treat everything as if it were a nail."}\\
\hfill-- {\scriptsize Abraham Harold Maslow, \emph{The Psychology of Science}~\cite{Maslow1966-MASTPO}} 
\end{quote}

Vision-Language-Action (VLA) models represent an emerging paradigm that extends foundation models to robotics by jointly modeling vision, language, and physical action spaces~\cite{kim2024openvla, zitkovich2023rt2, octo,Niu2024LLARVAVI}. While large-scale robotic datasets~\cite{openxembodiment2024rtx, khazatsky2024droid, dasari2019robonet} have enabled unprecedented training scales, VLAs have yet to achieve the impressive generalization of language and vision-language models. Unlike those models that demonstrate remarkable zero-shot adaptation, VLAs require targeted finetuning for each specific deployment environment, establishing a paradigm where practitioners must adapt models to match the exact specifications of their intended task. 

This reality raises a philosophical question: what constitutes a "task" in robotics?
A seemingly straightforward manipulation objective such as picking up a mug can have many physical instantiations when considering real-world perturbations~\cite{anwar2024contrast,taxonomy2025arxiv}, such as a camera position, the color of the mug, the table height, or even variations of the robot initial position by a few centimeters.
Unlike vision and language domains where tasks have clear boundaries, robotics operates in a continuous space of physical variations where the slightest environmental perturbation can fundamentally alter the required model behavior. 
We propose that few-shot expert demonstrations better specify what a robotic "task" is, as they contain the valuable physical information
inextricably linked to the task definition. Unlike linguistic descriptions alone, these demonstrations encode the physical properties of the deployment scenario: the exact angle the robot grasps from, how cluttered the workspace is, what lighting conditions exist, and countless other factors that determine successful execution.

Given task specification through few-shot demonstrations, the key challenge becomes: how can we effectively make use of these demonstrations to learn an embodied task efficiently? Current finetuning methods like LoRA~\cite{lora} adapt the \textit{same set of parameters regardless of the specific requirements of each task}.
In contrast, we take inspiration from functional specificity in neuroscience, which suggests that certain brain regions are specialized for particular tasks~\cite{kanwisher2000funcspec, FedorenkoFuncSpec}. Similarly, mechanistic interpretability in machine learning, which has shown that specific attention heads in transformers encode distinct capabilities~\cite{olsson2022context, hendel2023context, pmlr-v305-haon25a}. Building on these insights, we introduce a novel paradigm especially suited for robotics: using few-shot demonstrations, we first identify which attention heads encode task-relevant representations, and then we selectively finetune only those components. This approach recognizes that different tasks recruit different model capabilities, for example grasping from above requires different visual and spatial reasoning than pushing sideways, and adapts the model accordingly.

We introduce \textit{Robotic Steering}, the first approach to leverage mechanistic interpretability for finetuning task-specific representations of VLAs. Our method consists of three steps, each addressing a key challenge in VLA adaptation. First, we perform semantic attribution to identify task-relevant attention heads. Given a set of few-shot demonstrations of a task, we extract activations from each attention head as the base model performs a forward pass on the examples. We then select heads whose activations perform best on a lightweight k-NN regression task of predicting the ground truth actions for the examples. 
By identifying these task-specific heads, we can achieve more precise adaptation than uniformly finetuning all parameters.
Our second step is to freeze the visual encoder, action expert, and LLM backbone while applying targeted finetuning to only the queries and MLP parameters associated with selected heads using LoRA adapters. 
Finally, the resulting model deploys as a standard checkpoint without additional overhead. Unlike other mechanistic approaches that require activation interventions during inference, our finetuned weights integrate seamlessly into existing VLA deployment pipelines. 
An overview is shown in Figure~\ref{fig:teaser}, and a detailed view is shown in Figure~\ref{fig:main}.

We summarize the main contributions of our work: (i) We introduce Robotic Steering, the first method combining mechanistic interpretability with robotic finetuning for controllable adaptation through semantic attribution of attention heads; (ii) Through comprehensive on-robot evaluations using a Franka Emika robot arm, we demonstrate that Robotic Steering matches or outperforms full-head LoRA across all tested tasks while requiring less runtime and fewer parameters; (iii) Our approach exhibits superior task generalization and environmental robustness, including variations in lighting, object properties, and scene configurations, compared to standard finetuning methods; (iv) We provide a practical framework producing standard model checkpoints deployable without additional inference overhead, making mechanistic finetuning accessible for real-world robotic systems.

%% file: fig_2_main.tex
\begin{figure*}[t]
    \centering
    \includegraphics[width=.9\linewidth]{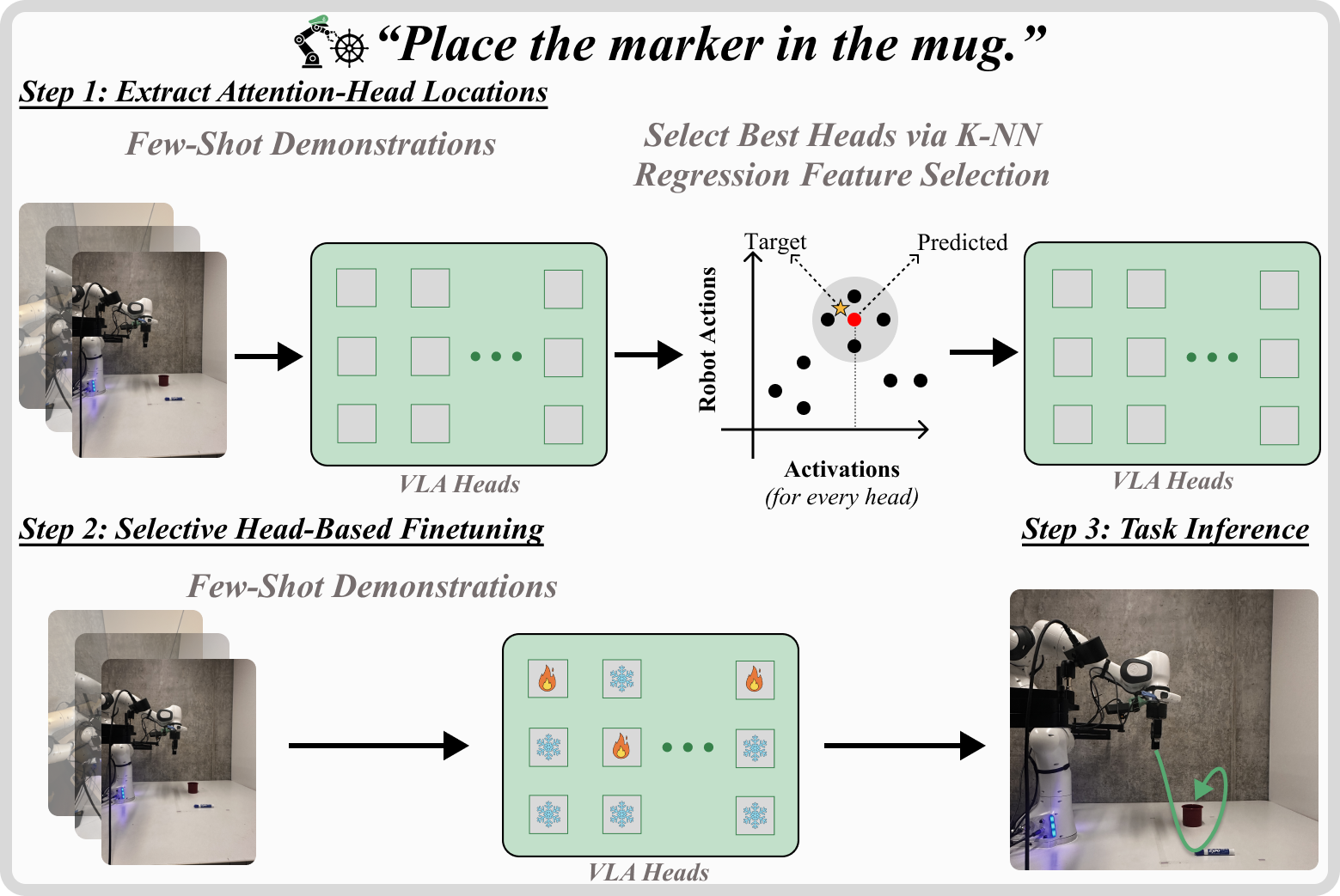}
    \caption{\textbf{Robotic Steering Approach.} Robotic Steering enables targeted adaptation of VLAs by (1) using few-shot demonstrations to extract task-relevant attention heads, (2) finetuning only these components, (3) and using the sparsely finetuned weights for task inference. }
    \label{fig:main}
\end{figure*}

%% file: sec/2_related_works.tex
\section{Related Work}

\minisection{Few-Shot Adaptation in Vision-Language-Action Models} Large Language Models (LLMs)~\cite{oggpt, Touvron2023LLaMAOA, bai2023qwen, Jiang2023Mistral7} and Large Multimodal Models (LMMs)~\cite{OpenAI2023GPT4TR, liu2023llava, liu2023llava15, Bai2023QwenVLAF, team2023gemini, Reid2024Gemini1.5, Anthropic2023Claude} have demonstrated remarkable capabilities through large-scale pretraining and causal token prediction. 
Vision-Language-Action models (VLAs) represent the current frontier of robot policy learning~\cite{reed2022generalist,driess2023palme,kim2024openvla,zitkovich2023rt2,octo} and is enabled by large-scale datasets~\cite{dasari2019robonet, openxembodiment2024rtx, khazatsky2024droid}.
This scale of training has demonstrated generalization across embodiments and tasks.
The state-of-the-art $\pi$-series models---$\pi_0$~\cite{black2024pi0} and $\pi_{0.5}$~\cite{physicalintelligence2025pi0.5}---use flow matching for continuous action generation along with large-scale data to achieve impressive zero-shot transfer. Despite these advances, VLAs struggle with few-shot adaptation to new environments.

Researchers have explored various few-shot techniques: in-context learning approaches~\cite{yin2025context, sridhar2025ricl, ma2024gvl} condition on demonstrations without weight updates but face context limitations; parameter-efficient methods~\cite{lora, Grover2025EnhancingGI, hu2024prompt, liu2023tail, liang2022transformer} and specialized adaptations~\cite{li2025controlvla, song2025few} reduce trainable parameters; meta-learning~\cite{finn2017oneshot, yu2018one, inproceedings} and behavior retrieval~\cite{du2023behavior,xie2025data,lin2024flowretrieval} enable rapid adaptation given access to prior data. However, these methods update parameters without considering which components encode task-specific physical reasoning, lacking interpretability and failing to leverage VLAs' structured representations. This motivates our mechanistic approach that identifies and selectively finetunes only task-relevant attention heads.

\minisection{Mechanistic Interpretability} Recent advances in mechanistic interpretability have revealed how model behavior can be precisely manipulated through internal representations. Early research~\cite{bau2017network, bau2020understanding, zhou2018interpreting} established frameworks for understanding semantic encoding in neural networks, while activation steering methods~\cite{subramani-etal-2022-extracting, Turner2024ActivationSteering, Panickssery2023CAA} demonstrated parameter-free behavior modification. The discovery of specialized components like induction heads~\cite{olsson2022context} and task-specific neurons~\cite{Hernandez2023Inspecting} led to task vector abstractions~\cite{hendel-etal-2023-context, Todd2024function}, with parallel work on sparse autoencoders~\cite{cunningham2023sae} and superposition~\cite{elhage2022superposition} providing tools for decomposing representations.

An emerging line of work leverages few-shot mechanistic interpretability for model adaptation via task vector methods~\cite{Hojel2024visual, huang2024multimodal, mitra2024sparse, chai2025activation}, which concentrate task-relevant information in specific attention heads or activation subspaces. Research in multimodal representations has revealed how vision-language models structure cross-modal concepts through multimodal neurons~\cite{goh2021multimodalneurons}, mechanistic understanding~\cite{schwettmann2023towards}, text-based decomposition~\cite{gandelsman2023textspan, balasubramanian2024decompose}, and knowledge localization~\cite{basu2024mloc, basu2024mlmstorage}. We are also excited by concurrent work on finetuning-free activation steering in VLAs that looks at controlling VLAs' behavior on in-domain tasks (e.g. controlling robot height and speed for a task)~\cite{pmlr-v305-haon25a}. The comprehensive survey by Lin et al.~\cite{lin2025mmfm_survey} provides a broader overview of such approaches. While these methods have succeeded in language and vision domains, our work is the first to apply mechanistic interpretability for \textit{finetuning} VLAs to learn new tasks in an efficient, interpretable, and robust manner.

%% file: tbl_1_main.tex
\begin{table*}[t]
\centering
\caption{\textbf{Results.} We report the performance and computational cost of \textbf{Robotic Steering} applied to real, on-robot tasks. Finetuned methods are trained on 20 demonstrations. All approaches are evaluated using 40 trials under the same task and environmental settings.}
\label{tab:main_results}
\small
\begin{tabularx}{\linewidth}{l*{7}{>{\centering\arraybackslash}X}}
\toprule
& \multicolumn{2}{c}{\textbf{Computational Cost}} & \multicolumn{5}{c}{\textbf{Tasks}} \\
\cmidrule(lr){2-3} \cmidrule(lr){4-8}
\textbf{Method} & \textbf{Training} & \textbf{Trainable} & \textbf{Place Marker} & \textbf{Press Button} & \textbf{Pick} & \textbf{Place Cube} & \textbf{Push Cup} \\
& \textbf{Time} & \textbf{Params} & \textbf{in Mug} & \textbf{Hard} & \textbf{Cube} & \textbf{in Bowl} & \textbf{to Bowl} \\
\midrule
\multicolumn{8}{l}{\textit{Zero-shot Methods}} \\
\midrule
$\pi_0$-DROID & - & - & 15\% & 0\% & 35\% & 5\% & 0\% \\
$\pi_{0.5}$-DROID & - & - & 10\% & 0\% & 20\% & 0\% & 0\% \\

\midrule
\multicolumn{8}{l}{\textit{Finetuned Methods}} \\
\midrule
$\pi_0$ Full-head LoRA & \textcolor{red}{239 min} & \textcolor{red}{1785.9M} & 75\% & 45\% & 75\% & 60\% & 10\% \\
\textbf{$\pi_0$ Robotic Steering (KNN)} & \textcolor{green!60!black}{189 min} & \textcolor{green!60!black}{78.8M} & \textbf{80\%} & \textbf{75\%} & \textbf{90\%} & \textbf{85\%} & \textbf{17.5\%} \\
\midrule
$\pi_{0.5}$ Full-head LoRA & \textcolor{red}{214 min} & \textcolor{red}{1781.3M} & 62.5\% & \textbf{90}\% & 70\% & 65\% & 20\% \\
\textbf{$\pi_{0.5}$ Robotic Steering (KNN)} & \textcolor{green!60!black}{185 min} & \textcolor{green!60!black}{74.6M} & \textbf{72.5}\% & 85\% & \textbf{77.5}\% & \textbf{80}\% & \textbf{27.5}\% \\

\bottomrule
\end{tabularx}
\end{table*}

%% file: sec/3_methods.tex
\section{Methods}
\label{sec:methods}

In this section, we present Robotic Steering, a finetuning approach inspired by mechanistic interpretability that updates task-specific components of Vision-Language-Action models. Our method identifies and selectively finetunes attention heads that encode task-relevant physical reasoning, allowing VLAs to learn new capabilities and preserve existing ones. We begin with preliminaries on VLA architectures, followed by our three-step approach: (1) identifying task-relevant attention heads, (2) selective finetuning of identified components, and (3) standard inference with finetuned weights.

\subsection{Preliminaries}
\label{sec:methods:preliminaries}

\minisection{Vision-Language-Action Models}
VLAs extend the transformer architecture to robotic control by processing visual observations and language instructions to predict continuous action vectors. Given an observation $o_t$ consisting of image frames and optional language instruction, a VLA predicts an action vector $a_t \in \mathbb{R}^d$ containing control values (e.g., joint velocities, gripper commands). Modern VLAs like $\pi_0$ and $\pi_{0.5}$~\cite{black2024pi0, physicalintelligence2025pi0.5} formulate this as a conditional generation problem, where actions are produced through autoregressive token prediction or flow matching. The model processes inputs as a sequence of visual tokens, language tokens, and robot state information, conditioning on this multimodal information for action prediction.

\minisection{Multi-Head Attention}
For a transformer with $L$ layers and $H$ attention heads per layer, each head $(l, h)$ computes:
\begin{equation}
\mathbf{h}_l^h(x_i) = \text{softmax}\left(\frac{QK^T}{\sqrt{d_h}}\right)V
\end{equation}
where $Q$, $K$, $V$ are query, key, and value projections. For action prediction in VLAs, we focus on activations at the final token position $\mathbf{h}_l^h(x_T)$, which aggregates information across the entire sequence.
\input{tbl_2_generalization}

\subsection{Step 1: Identifying Task-Relevant Attention Heads}
\label{sec:methods:head_selection}

Our key insight is that within a VLA's attention mechanism, specific heads naturally specialize in encoding physical concepts relevant to particular manipulation tasks. We identify these heads through their ability to retrieve examples with similar action patterns.

\minisection{Extracting Head Activations}
Suppose we are given a frozen VLA and few-shot demonstrations $\mathcal{D} = \{(\tau_1, a_1), (\tau_2, a_2), \dots, (\tau_N, a_N)\}$, where each trajectory $\tau_i$ consists of $T$ timesteps. Each timestep $t$ contains the VLA's input observation: visual tokens from camera images, language tokens from task instructions, and robot state information (e.g., joint angles). The corresponding $a_i \in \mathbb{R}^{T \times d}$ are action vectors across all timesteps.

For each timestep $t$ in trajectory $\tau_i$, we extract the attention vector $\mathbf{h}_l^h(\tau_i^t)$ for every head $(l, h)$. Importantly, we work at the timestep level rather than trajectory level---each timestep becomes an individual example in our retrieval set.

\minisection{k-NN Regression for Head Evaluation}
\label{sec:knn_regression}
To evaluate each head's relevance, we assess its ability to retrieve timesteps with similar actions. The intuition is that if a head's representation groups together observations that require similar physical actions, then this head encodes task-relevant features worth finetuning. In order to make head selection more efficient, we employ the keyframe extraction approach suggested in~\cite{yin2025context}. Functionally, however, the approach is identical with or without this step. More details can be found in Section~\ref{supp:additional} of the Supplementary material.

For a query observation $q$ from trajectory $\tau_i$ at timestep $t$:

We first find the $k$ nearest neighbor timesteps from all other trajectories based on cosine similarity in head $(l, h)$'s representation space:
\begin{equation}
\mathcal{N}_k^{l,h}(q) = \text{top-}k\left\{\frac{\mathbf{h}_l^h(q) \cdot \mathbf{h}_l^h(\tau_j^s)}{\|\mathbf{h}_l^h(q)\| \|\mathbf{h}_l^h(\tau_j^s)\|}\right\}_{j \neq i, s}
\end{equation}
Second, we predict the action by averaging the actions of retrieved neighbors:
\begin{equation}
\hat{a}^{l,h}_t = \frac{1}{k} \sum_{(\tau_j^s) \in \mathcal{N}_k^{l,h}(q)} a_j^s
\end{equation}
Finally, we compute the head's score as the mean squared error across all queries:
\begin{equation}
\text{score}(l, h) = \frac{1}{|\mathcal{D}| \cdot T} \sum_{\tau_i \in \mathcal{D}} \sum_{t=1}^T \|\hat{a}^{l,h}_t - a_i^t\|^2
\end{equation}
We select the top-$m$ heads with lowest scores:
\begin{equation}
\mathcal{H}_{\text{task}} = \{(l,h) \mid \text{score}(l,h) \text{ is among } m \text{ lowest scores}\}
\end{equation}
We select $k$ empirically by evaluating different values against the MSE criterion (details in Section~\ref{supp:additional} of the Supp.).

These heads learn representations that effectively map task-specific observations to other observations in a few-shot demonstration set that require similar actions, making them ideal candidates for task-specific finetuning.

\subsection{Step 2: Selective Finetuning with LoRA}
\label{sec:methods:finetuning}

Having identified task-relevant heads $\mathcal{H}_{\text{task}}$, we perform targeted finetuning on those particular parameters.

\minisection{Sparse Parameter Updates}
We freeze all model components except the query projections of selected heads. For each head $(l,h) \in \mathcal{H}_{\text{task}}$, we apply Low-Rank Adaptation (LoRA)~\cite{lora}:
\begin{equation}
W_Q'^{l,h} = W_Q^{l,h} + B^{l,h}A^{l,h}
\end{equation}
where $B^{l,h} \in \mathbb{R}^{d \times r}$ and $A^{l,h} \in \mathbb{R}^{r \times d}$ are low-rank matrices with rank $r \ll d$. We also finetune the MLP layers associated with the selected attention blocks.

\minisection{Training Objective}
Our approach is flexible and compatible with any VLA training objective. We simply finetune the selected heads using the same loss function as the base model---whether that's flow matching loss for diffusion-based models like $\pi_{0.5}$ or cross-entropy for discretized action spaces. This selective updating acts as a targeted refinement that enhances task performance without broadly overwriting all of the model's parameters.
\input{tbl_3_ablations}
\subsection{Step 3: Inference}
\label{sec:methods:inference}

After selective finetuning, inference proceeds through standard forward passes with the finetuned weights. Unlike many mechanistic interpretability methods that require computing and manipulating activations at inference time, our approach produces a standard model checkpoint deployable without additional computational overhead or specialized procedures. The model simply uses the finetuned weights for the selected heads while maintaining frozen weights elsewhere, preserving both new task capabilities and existing skills.

%% file: tbl_2_generalization.tex
\begin{table*}[t]
\centering
\caption{\textbf{Generalization and Robustness.} We compare \textbf{Robotic Steering} to full-head LoRA in both single and multi-task settings. We also compare both methods' generalization to unseen tasks and robustness to environmental variability. All methods are finetuned on 20 demonstrations \textit{per task} and evaluated on 40 trials. Results are shown for $\pi_{0.5}$ VLA model.}
\label{tab:multitask_results}
\small
\begin{tabularx}{\linewidth}{l*{7}{>{\centering\arraybackslash}X}}
\toprule
& & \multicolumn{3}{c}{\textbf{Task Generalization}} & \multicolumn{3}{c}{\textbf{Task Robustness}} \\
\cmidrule(lr){3-5} \cmidrule(lr){6-8}
\textbf{Method} & \textbf{Training} & \textbf{Place Marker} & \textbf{Place Cube} & \textbf{Pick Mug} & \textbf{Lighting} & \textbf{Form} & \textbf{Distractor Object} \\
& \textbf{Strategy} & \textbf{in Mug} & \textbf{in Bowl} & \textbf{(Unseen)} & \textbf{Variation} & \textbf{Variation} & \textbf{Variation} \\
\midrule
\multicolumn{8}{l}{\textit{Single-Task Training (Place Marker in Mug only)}} \\
\midrule
LoRA & Single & 62.5\% & 20\% & 42.5\% & 25\% \textcolor{red}{$\downarrow$37.5\%} & 22.5\% \textcolor{red}{$\downarrow$40\%} & 30\% \textcolor{red}{$\downarrow$32.5\%} \\
Robotic Steering & Single & \textbf{72.5\%} & \textbf{25\%} & \textbf{65\%} & \textbf{47.5\%} \textcolor{orange}{$\downarrow$25\%} & \textbf{37.5\%} \textcolor{orange}{$\downarrow$35\%} & \textbf{40\%} \textcolor{orange}{$\downarrow$32.5\%} \\
\midrule
\multicolumn{8}{l}{\textit{Multi-Task Training (Place Marker + Place Cube)}} \\
\midrule
LoRA & Joint & 37.5\% & 37.5\% & 0\% & 15\% & 12.5\% & 15\% \\
Robotic Steering & Non-overlapping & \textbf{45\%} & \textbf{65\%} & 12.5\% & \textbf{30\%} & 17.5\% & 25\% \\
Robotic Steering & Joint Selection & 40\% & 47.5\% & \textbf{20\%} & 22.5\% & \textbf{20\%} & \textbf{27.5\%} \\
\bottomrule
\end{tabularx}
\end{table*}

%% file: tbl_3_ablations.tex
\begin{table*}[t]
\centering
\caption{\textbf{Ablations.} We compare the performance and cost of different design choices for head-selection and training configuration.  All methods use 20 few-shot demos and select top-20 heads for adaptation.}
\label{tab:ablations}
\small
\begin{tabularx}{\linewidth}{l*{5}{>{\centering\arraybackslash}X}}
\toprule
\textbf{Method} & \textbf{Place Marker} & \textbf{Head Selection Time} & \textbf{Fine-tuning Time} & \textbf{Activation Cache} & \textbf{VLA Model} \\
& \textbf{in Mug} & \textbf{(min)} & \textbf{(min)} & \textbf{(M)} & \textbf{Inference}\\
\midrule
\multicolumn{4}{l}{\textit{Head Selection Methods}} \\
\midrule
CMA & 15\% & 58 min & 188 min & 92.83M & Required\\
REINFORCE & 80\% & 93 min & 186 min & 92.83M & Required\\
\textbf{K-NN Regression (Ours)} & \textbf{80\%} & \textbf{0.2 min} & 185 min & 92.83M & \textbf{Not Needed}\\
\midrule
\multicolumn{5}{l}{\textit{Training Components}} \\
\midrule
Queries only & 10\% & 0.2 min & 186 min & 92.83M & - \\
\textbf{Queries + MLP (Ours)} & \textbf{80\%} & \textbf{0.2 min} & \textbf{185 min} & 92.83M & - \\
\bottomrule
\end{tabularx}
\end{table*}

%% file: sec/4_setup_details.tex
\section{Evaluation}
\label{sec:evaluation}

In our work, we evaluate our method on a variety of real-world on-robot tasks using the strong $\pi_0$ and $\pi_{0.5}$ VLAs to demonstrate the effectiveness of our approach on realistic, physically-grounded usecases. We select tasks of diverse difficulties and skills and deeper experimentation and ablation that showcases the many unique qualities of our approach including its performance, robustness, and interpretability. We present more details as follows:

\subsection{Implementation Details}
\label{sec:implementation}

While our method is model-agnostic, we use $\pi_0$~\cite{black2024pi0} and $\pi_{0.5}$~\cite{physicalintelligence2025pi0.5}, two state-of-the-art VLAs that use flow matching for continuous action generation. Our entire implementation is in Jax~\cite{jax2018github}, which notably lacks convenient hooks to easily extract activations from the model. Thus, we highlight the development of such functionality for a Jax-based model as a core technical contribution of our work. We finetune the model using 2 NVIDIA RTX A6000 GPUs, emphasizing the lightweight nature of our approach. We extract attention activations from the model's PaliGemma~\cite{beyer2024paligemma, steiner2024paligemma} LLM backbone with 18 layers with 8 heads each, selecting $m=20$ heads for finetuning based on k-NN regression with $k \in \{10, 20, 30, 40\}$ neighbors and choosing the $k$ that yields the lowest mean squared error. The LoRA rank is set to $r=8$, and we finetune for 5,000 steps for our main experiments using only 20 demonstrations of the target task. More implementation details are in Supp. Section~\ref{supp:implementation}. 

\subsection{Robotic Setup Details}
\label{sec:robot_setup}
We follow the setup from DROID~\cite{khazatsky2024droid} exactly, using a 7-DoF Franka Emika Panda robot arm with a Robotiq gripper and a low-level Polymetis controller~\cite{Polymetis2021}. As suggested by DROID, we enable two of the three cameras for both finetuning and inference: the left arm camera and wrist camera. We record each example episode at 6 Hz. All data collection is performed on-robot using teleoperation, with each task controlling for the exact objects used to ensure fair evaluation across methods.

We evaluate a total of 5 primary tasks with the following language instructions: (1) "place marker in mug", (2) "press red button hard", (3) "pick up red cube", (4) "place green cube in red bowl", and (5) "push red cup to red bowl". We collect 20 teleoperated demonstrations for all tasks for both head selection and finetuning, representing a sample-efficient few-shot paradigm. All models are finetuned for 5,000 iterations as detailed in Section~\ref{sec:implementation} (implementation details).
More details about the robot and the task setup can be found in Section~\ref{supp:robot_setup} of the Supplement.


%% file: sec/5_experiments_and_results.tex
\section{Results}
\input{fig_3_scaling}
Our main results are shown in Table~\ref{tab:main_results}. The crucial insight of Robotic Steering is that few-shot expert demonstrations can encode the physical nuances of robotic tasks and more importantly inform which task-specific components of a model to finetune for model adaptation.

Indeed, our results demonstrate that Robotic Steering matches or outperforms LoRA's success rate on all evaluated tasks across both $\pi_0$ and $\pi_{0.5}$ VLA models. This holds true for both simpler in-domain tasks which are similar to DROID~\cite{khazatsky2024droid} dataset tasks and more challenging new tasks. This demonstrates that Robotic Steering is a broadly effective finetuning approach, leveraging only 20 demonstrations for both head selection and finetuning to surpass the LoRA baseline. It is worth noting that none of these tasks are trivial given the physical context of on-robot evaluation, as evidenced by near 0\% zero-shot success rates for most tasks.

Interestingly, $\pi_0$ and $\pi_{0.5}$ exhibit similar performance on these short-horizon manipulation tasks. We hypothesize this similarity stems from two factors: first, $\pi_0$ already achieves strong performance on this task distribution, leaving limited room for improvement; and second, $\pi_{0.5}$'s methodological improvements over $\pi{0}$ are significantly centered around long-horizon reasoning tasks, wheras our tasks are shor-horizon for simpler and more consistent comparison. 

Beyond task performance, Table~\ref{tab:main_results} demonstrates that Robotic Steering is significantly more computationally efficient than full-head LoRA, reducing finetuning time by 21\% while using 96\% fewer trainable parameters. This efficiency is crucial for practical robotics, where rapid iteration and experimentation in new environments is essential.

Additional results and ablations are in Section~\ref{supp:experiments} in Supp.

\subsection{Ablations}
We perform a comprehensive ablation study of Robotic Steering on the \textit{Place Marker in Mug} task to understand the impact of key design choices. For all ablations, we use $\pi_{0.5}$.

\minisection{Scaling number of demonstrations} 
In Figure~\ref{fig:scaling_demos}, we analyze the effect of varying the number of demonstrations used for fine-tuning. Robotic Steering consistently outperforms Full-head LoRA across all data scales, achieving higher success rates especially in low-data regimes. Performance improves sharply up to 20 demonstrations, reaching around 72.5\% success rate for Robotic Steering compared to 62.5\% for Full-head LoRA, and then saturates. This demonstrates that our method can make more efficient use of limited demonstrations while maintaining superior scalability.

\minisection{Scaling with training iterations} 
We investigate how our method scales with the number of training iterations compared to Full-head LoRA. As shown in Figure~\ref{fig:scaling_iterations}, Robotic Steering demonstrates faster initial learning and achieves higher final performance (72.5\%) compared to Full-head LoRA (62.5\%) after 5k iterations. This result suggests that our approach surpasses or at least matching LoRA's capabilities of scaling performance with further training.

\minisection{Head selection approach}
Our results in Table~\ref{tab:ablations} show that K-NN regression, our approach for head selection, slightly outperforms Causal Mediation Analysis (CMA)~\cite{Todd2024function} and REINFORCE~\cite{Hojel2024FindingVT, huang2024multimodal}. CMA, more specifically implemented as causal ablation in our experiments, selects heads by adding noise to each head and measuring the resulting performance drop on the 20 few-shot demonstrations. REINFORCE optimizes head selection through gradient-based search to maximize task performance. While all three methods achieve comparable task success rates as shown in Table~\ref{tab:ablations}, K-NN regression offers a crucial advantage: significantly lower runtime. This is due to K-NN regression not requiring model inference and evaluation for head selection. Once the activations are computed, K-NN regression becomes a simple and efficient regression approach on the activations themselves. 

\minisection{Training Components}
We also carefully ablate the recipe for which precise components of the model to finetune. Of course, when selecting heads, it is natural to finetune their queries, but we also question whether additionally finetuning their MLPs, yields any benefit. Our results in Table~\ref{tab:ablations} suggest that indeed finetuning both the queries and MLPs associated with the selected task-specific heads yields improvements in success rate. This suggests that the feedforward projection following attention is important to adapt for VLA finetuning. We do not consider finetuning the parameters of the keys and values as they are shared per layer in $\pi_0$'s base LLM~\cite{black2024pi0}.

\input{fig_4_head_vis}
\subsection{Additional Experiments}
\label{sec:add_expr}

A key strength of Robotic Steering is its flexibility and generalization across different training scenarios. In this section, we demonstrate that our approach not only improves task-specific performance but also exhibits robust generalization capabilities in three important dimensions: multi-task learning, transfer to unseen tasks, and robustness to environmental variations. All experiments use $\pi_{0.5}$. Results are shown in Table~\ref{tab:multitask_results}, Figure~\ref{fig:scaling_experiments}, and Section~\ref{supp:additional} of the Supplementary.

\minisection{Multi-task training strategies}
Because Robotic Steering selects task-specific heads, our approach naturally accommodates multiple strategies for multi-task learning. We explore two complementary approaches. \textit{Non-overlapping head selection} trains each task on a distinct set of heads, where k-NN regression independently selects the 20 most task-relevant heads for each individual task. This allows tasks to specialize to their own optimal head subset. \textit{Joint head selection} uses a unified set of examples and activations from both tasks to select a single set of 20 heads optimized for joint performance, enabling shared representations across the task pair. Across both multi-task approaches, Robotic Steering outperforms full-head LoRA consistently across both trained tasks suggesting that Robotic Steering works well as a multi-task learning framework.

\minisection{Generalization to unseen tasks}
While trained exclusively on Place Marker in Mug and Place Cube in Bowl, our method generalizes successfully to an unseen task (Pick Mug). Excitingly, both settings of multi-task finetuning lead to noticeable gains over LoRA on the unseen task. This result shows us that while Robotic Steering leads to superior performance on the specific tasks in the train set, it also offers broader generalization to similar tasks, a capability that LoRA appears to lack.

\minisection{Robustness to environmental variations}
A critical concern in real-world robotics is robustness to environmental changes that occur during deployment. We evaluate both Robotic Steering and LoRA under three types of perturbations to the Place Marker in Mug task: lighting variation, form variation (e.g. different marker shapes), and the introduction of distractor objects. Importantly, Robotic Steering demonstrates significantly greater robustness across all three conditions compared to full-head LoRA. This suggests that task-specific sparse head selection naturally filters out task-irrelevant features while preserving robust, task-critical representations. This result is particularly exciting as it challenges a reasonable assumption that sparse, task-specific parameter selection might be brittle; instead, our approach demonstrates that mechanistic steering produces generalizable and robust finetuning even under distributional shift.

\minisection{Interpreting task-specific attention heads}
To understand what our method learns, we visualize the attention patterns of the top-selected heads for Place Marker in Mug and Place Cube in Bowl in Figure~\ref{fig:head_viz}. The visualizations confirm that different tasks activate distinct sets of heads, consistent with the principle of functional specificity in attention mechanisms. This interpretability can be a fundamental advantage of Robotic Steering: unlike general adaptation methods, we can directly inspect and understand which attention mechanisms are being leveraged for each task. We encourage future works to explore the interpretability of these task representations further. Visualizations of more tasks can be found in Section~\ref{supp:additional visualization} of the Supplement.

%% file: fig_3_scaling.tex

\begin{figure*}[t]
\centering
\begin{subfigure}[b]{0.48\textwidth}
  \centering
  \includegraphics[width=\textwidth]{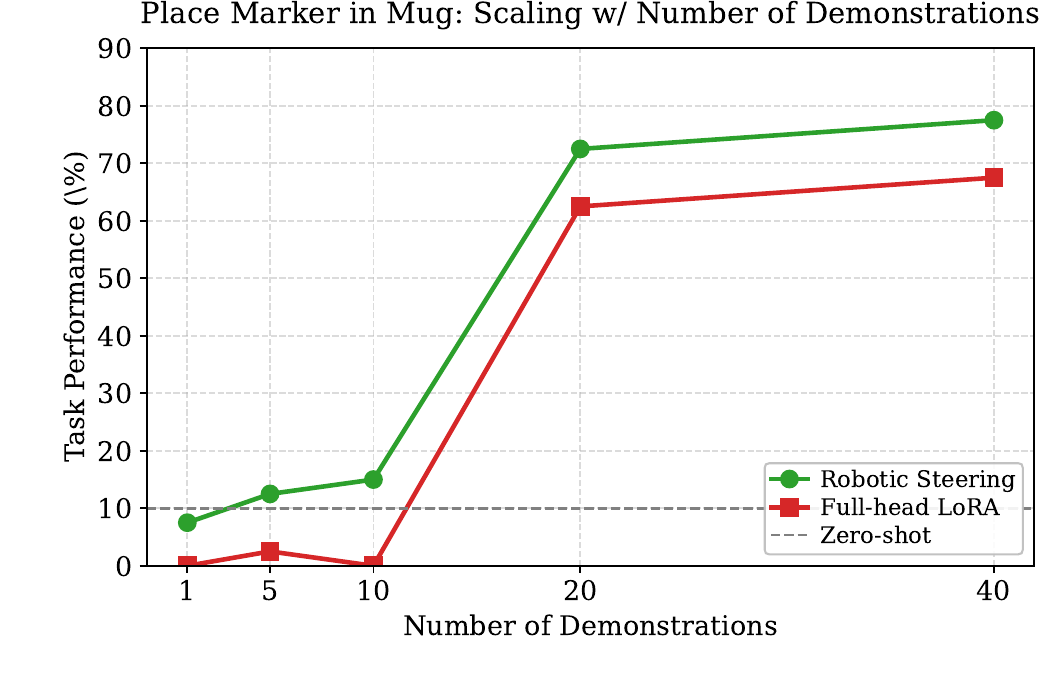}
  \caption{Number of Demonstrations}
  \label{fig:scaling_demos}
\end{subfigure}
\hfill
\begin{subfigure}[b]{0.48\textwidth}
  \centering
  \includegraphics[width=\textwidth]{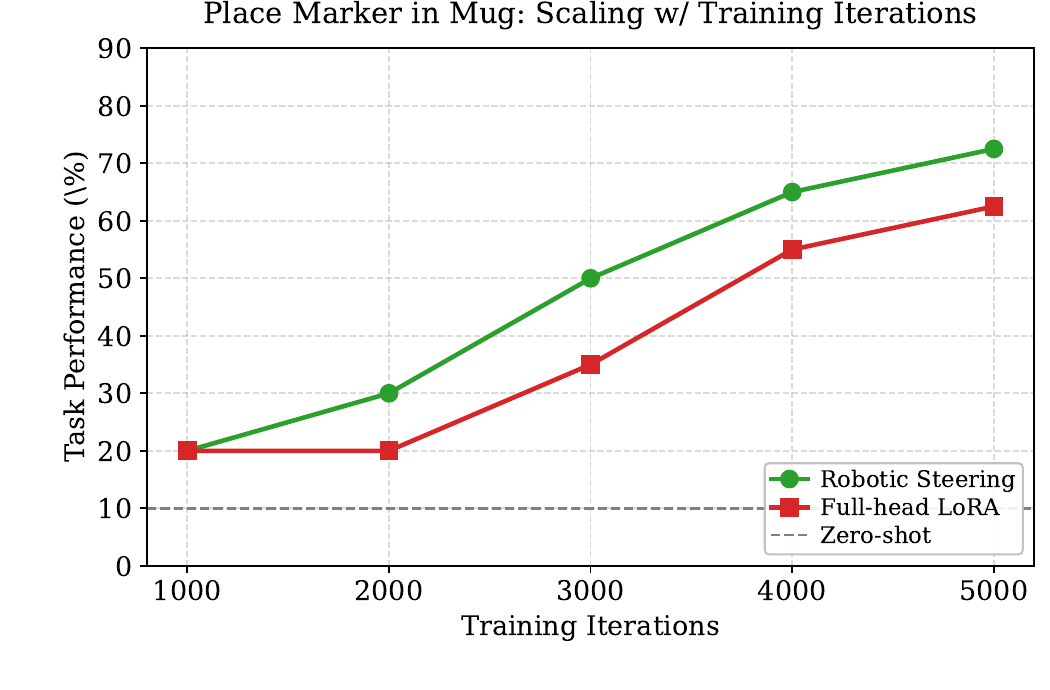}
  \caption{Training iterations}
  \label{fig:scaling_iterations}
\end{subfigure}
\caption{\textbf{Scaling Experiments.} For the \textit{Place Marker in Mug} task, we show the (a) success rate versus number of demonstrations and (b) success rate versus number of training iterations for Robotic Steering and full-head LoRA}
\label{fig:scaling_experiments}
\end{figure*}


%% file: fig_4_head_vis.tex
\begin{figure*}
    \centering
    \includegraphics[width=.9\linewidth]{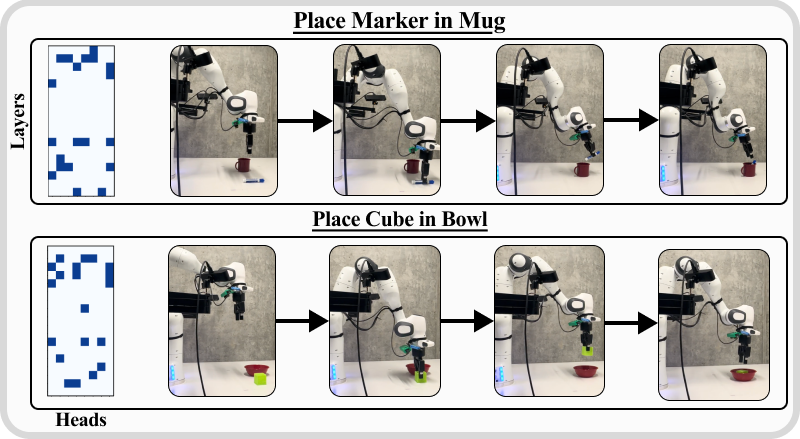}
    \caption{\textbf{Task and Attention Head Selection Visual.} We show the selected heads and task visuals for Robotic Steering on the  \textit{Place Marker in Mug} task (top) and the \textit{Place Cube in Bowl} task (bottom).}
    \label{fig:head_viz}
\end{figure*}

%% file: sec/6_conclusion_and_limitations.tex
\section{Conclusion}
In this work, we introduce Robotic Steering, which demonstrates that few-shot demonstrations can specify physically-grounded embodied tasks and help identify which attention heads in VLAs encode task-relevant physical reasoning. By selectively finetuning these heads, we match or exceed full LoRA performance while using 96\% fewer parameters, generalizing to unseen tasks, and being robust to environmental variations. Our visualizations reveal that different manipulation tasks activate distinct attention patterns, providing mechanistic insight into how VLAs encode physical tasks.

This work opens exciting research directions at the intersection of mechanistic interpretability and robotic learning. Future methods could explore alternative head selection approaches beyond K-NN regression, investigate parameter-level feature-selection approaches, and consider applications to long-horizon tasks. More fundamentally, our results suggest that the question of ``what to finetune" deserves equal attention as      ``how to finetune", a shift that could transform how we adapt foundation models for robotics. As VLAs scale to billions of parameters, the ability to precisely identify and modify task-relevant components will become essential for practical deployment across the wide variety of physical contexts robots must learn to master.

\section{Limitations}
We outline a few limitations of our work. Firstly, Robotic Steering requires open-source access to the weights which may not be possible on closed-source models. Secondly, although our approach is completely embodiment agnostic, we evaluate on a single Franka Emika Panda robot arm due to resource constraints. We encourage the application of our work to a wide variety of embodiments and environments. Lastly, our interpretable feature extraction focuses on attention heads of the LLM, when there can be informative features in the other model components as well. We greatly encourage future research in further exploring the intersection of mechanistic interpretability and robotics.

%% file: sec/7_supp.tex
\maketitlesupplementary
\newpage
\appendix

Here, we provide additional details on experiments and ablation studies (\Secref{supp:experiments}), implementation details (\Secref{supp:implementation}), and the robotic hardware setup (\Secref{supp:robot_setup}).

\section{Additional Experimental Results}
\label{supp:experiments}

\subsection{Ablation Studies}
\label{supp:ablations}

\minisection{Token Position Selection for Head Activations} To determine the optimal token position for attention head selection, we evaluated activations of 20 demos on the task place marker in mug from two positions using our k-NN regression method (see Section~\ref{sec:knn_regression}): the \textit{last token in prefix} and the \textit{first token in suffix}. We tested both positions under $\pi_0$ and $\pi_{0.5}$ architectures. We use the Coefficient of Variation (CV = $\frac{\sigma}{\mu}$) to measure head selection distinctiveness, where higher CV indicates greater variance in head informativeness. As shown in Table~\ref{tab:token_ablate}, the first token in suffix consistently exhibits higher CV values than the last token in prefix. For $\pi_0$, the first token achieves CV of 0.12 versus 0.04 for the last token. For $\pi_{0.5}$, the values are 0.47 versus 0.05, respectively. Based on these results, we select the first token in suffix for attention head analysis and selection in both architectures for all subsequent experiments.



\minisection{Distance Metric for k-NN Regression} There are many choices of distance metrics for KNN-based feature comparison. Our features are activations extracted from the state token, which attends to all preceding tokens (i.e., vision and language tokens). Consequently, these activations encode visual, language, and robot-state information. We compared cosine similarity and Euclidean distance as the KNN metric, and found that heads selected using cosine similarity has lower MSE with respect to the ground-truth actions than those selected with Euclidean distance. We therefore choose cosine similarity as the default metric.

\minisection{Number of Nearest Neighbors (k)} The value of $k$ in our k-NN regression method (see Section~\ref{sec:knn_regression}) is not fixed but optimized per task. For each task, we search over $k \in \{10, 20, 30, 40\}$ and select the top 20 heads based on each candidate $k$ value. We then evaluate the MSE performance of each resulting head subset and choose the $k$ that has the lowest MSE loss.

\begin{figure}[t]
\centering
\includegraphics[width=0.9\columnwidth]{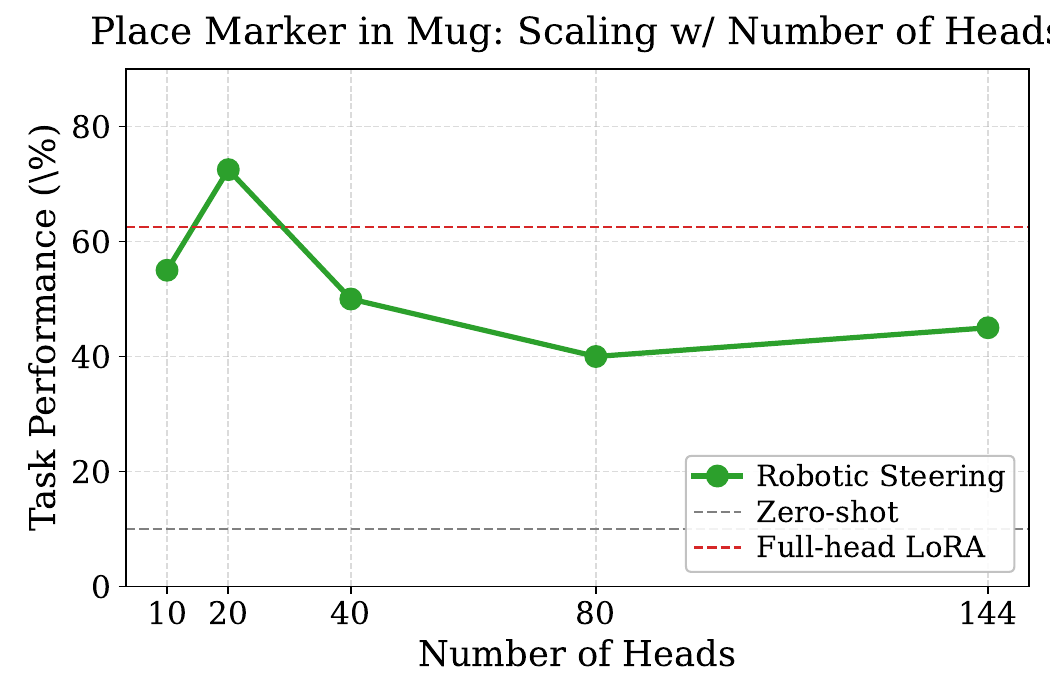}
\caption{\textbf{Scaling Experiments.} For the \textit{Place Marker in Mug} task, we show the success rate versus number of heads selected for fine-tuning.}
\label{fig:scaling_head}
\end{figure}

\minisection{Number of Heads Fine-tuned}
\label{supp:ablations:num_heads}
We ablated the number of heads selected for fine-tuning on the \textit{Place Marker in Mug} task. As shown in Figure~\ref{fig:scaling_head}, performance peaks at 20 heads with 72.5\% success rate, outperforming both zero-shot (10\%) and full-head LoRA (62.5\%). Performance degrades with fewer heads or more heads , suggesting that too many heads introduce task-irrelevant information.

Notably, fine-tuning all 144 heads with Robotic Steering achieves only 45\%, lower than full-head LoRA's 62.5\%. This gap occurs because \textit{Robotic Steering} additionally freezes the vision encoder and action expert, while full-head LoRA finetuning updates all parameters.


\begin{table}[!t]
\centering
\caption{Head Selection Distinctiveness across Different Token Positions.}
\label{tab:token_ablate}
\footnotesize
\begin{tabular}{lcccc}
\toprule
\multirow{2}{*}{\textbf{Metric}} & \multicolumn{2}{c}{\textbf{Last Token in Prefix}} & \multicolumn{2}{c}{\textbf{First Token in Suffix}} \\
\cmidrule(lr){2-3} \cmidrule(lr){4-5}
& $\pi_0$ & $\pi_{0.5}$ & $\pi_0$ & $\pi_{0.5}$ \\
\midrule
Coefficient of Variation & 0.04 & 0.05 & \textbf{0.12} & \textbf{0.47} \\
\bottomrule
\end{tabular}
\end{table}

\subsection{Additional Experiments}
\label{supp:additional}

\minisection{Full Fine-tuning} Besides LoRA fine-tuning, we also evaluated how our method works in full fine-tuning. 2 NVIDIA RTX A100 GPUs are used for full-fine-tuning. We can see from Table~\ref{tab:FFT_results}, \textit{Robotic Steering} still outperforms simple fine-tuning approach in full fine-tuning setting.

\begin{table*}[t]
\centering
\caption{\textbf{Results.} Performance and computational cost of \textit{Robotic Steering} in Full Fine-tuning.}
\label{tab:FFT_results}
\small
\begin{tabularx}{\linewidth}{l*{5}{>{\centering\arraybackslash}X}}
\toprule
& \multicolumn{2}{c}{\textbf{Computational Cost}} & \multicolumn{3}{c}{\textbf{Tasks}} \\
\cmidrule(lr){2-3} \cmidrule(lr){4-6}
\textbf{Method} & \textbf{Training} & \textbf{Trainable} & \textbf{Place Marker} & \textbf{Press Button} & \textbf{Place Cube} \\
& \textbf{Time} & \textbf{Params} & \textbf{in Mug} & \textbf{Hard} & \textbf{in Bowl} \\
\midrule
\multicolumn{6}{l}{\textit{Zero-shot Methods}} \\
\midrule
$\pi_0$-DROID & - & - & 15\% & 0\% & 5\% \\
$\pi_{0.5}$-DROID & - & - & 10\% & 0\% & 0\% \\
\midrule
\multicolumn{6}{l}{\textit{Finetuned Methods}} \\
\midrule
$\pi_{0.5}$ Full-head Full Fine-tuning & \textcolor{red}{110 min} & \textcolor{red}{12799.5M} & 45\% & 77.5\% & 55\% \\
\textbf{$\pi_{0.5}$ Robotic Steering (KNN)} & \textcolor{green!60!black}{85 min} & \textcolor{green!60!black}{9012.5M} & \textbf{60}\% & \textbf{87.5}\% & \textbf{72.5}\% \\
\bottomrule
\end{tabularx}
\end{table*}

\minisection{Consistency of Head Selection Across Data Subsets} To evaluate the robustness of our head selection method, we analyzed consistency across 5 data variants of the \textit{Place Marker in Mug} task: (1) all 40 demonstrations, (2) 20 randomly sampled demos with seed 42, (3) 20 randomly sampled demos with seed 24, (4) the first 20 demos, and (5) the last 20 demos. We applied our k-NN regression method to select the top-20 and top-40 heads for each variant.

\begin{figure*}[t]
\centering
\begin{subfigure}[b]{0.48\textwidth}
  \centering
  \includegraphics[width=\textwidth]{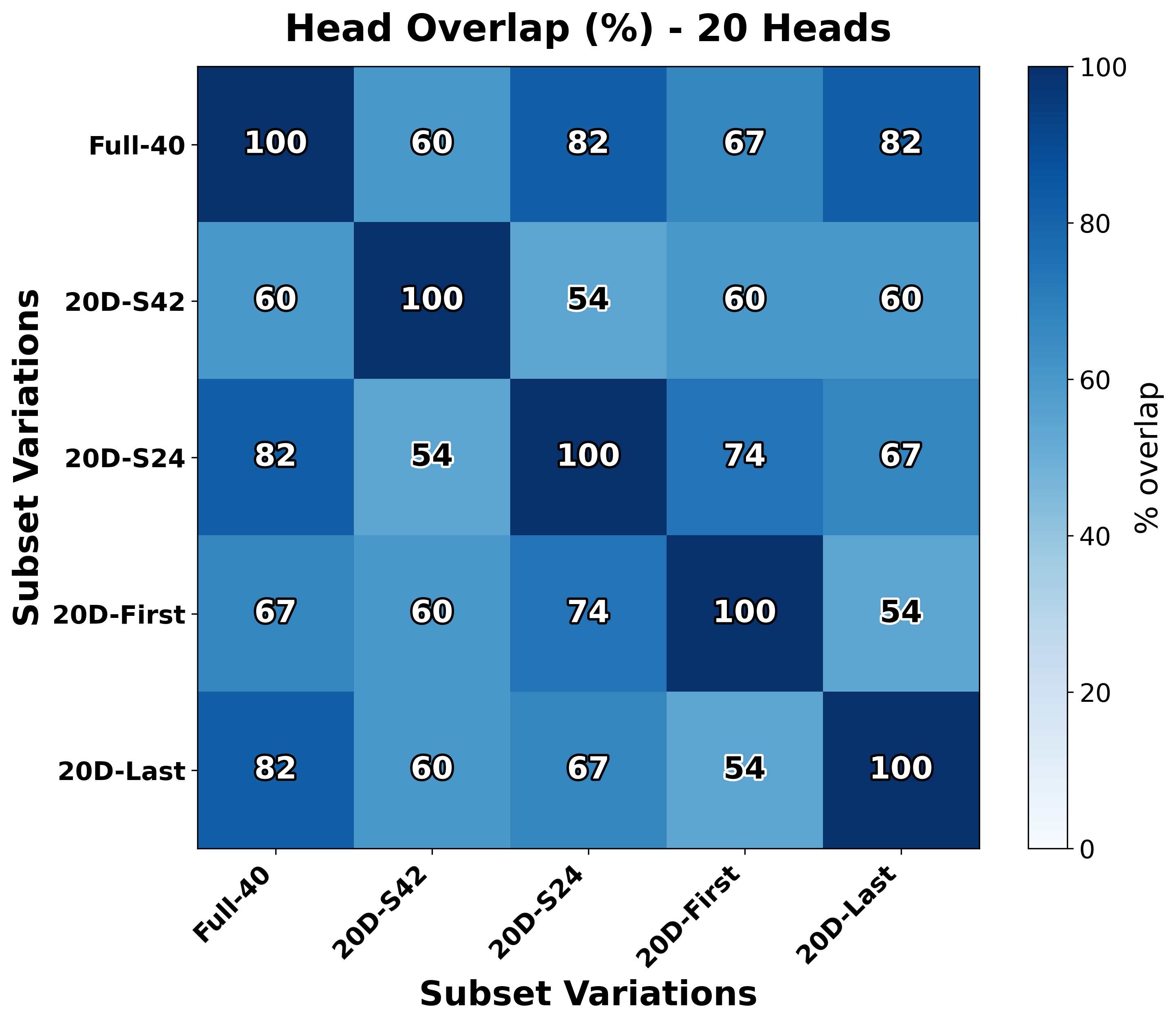}
  \caption{Top-20 Head Selection}
  \label{fig:head_overlap_20}
\end{subfigure}
\hfill
\begin{subfigure}[b]{0.48\textwidth}
  \centering
  \includegraphics[width=\textwidth]{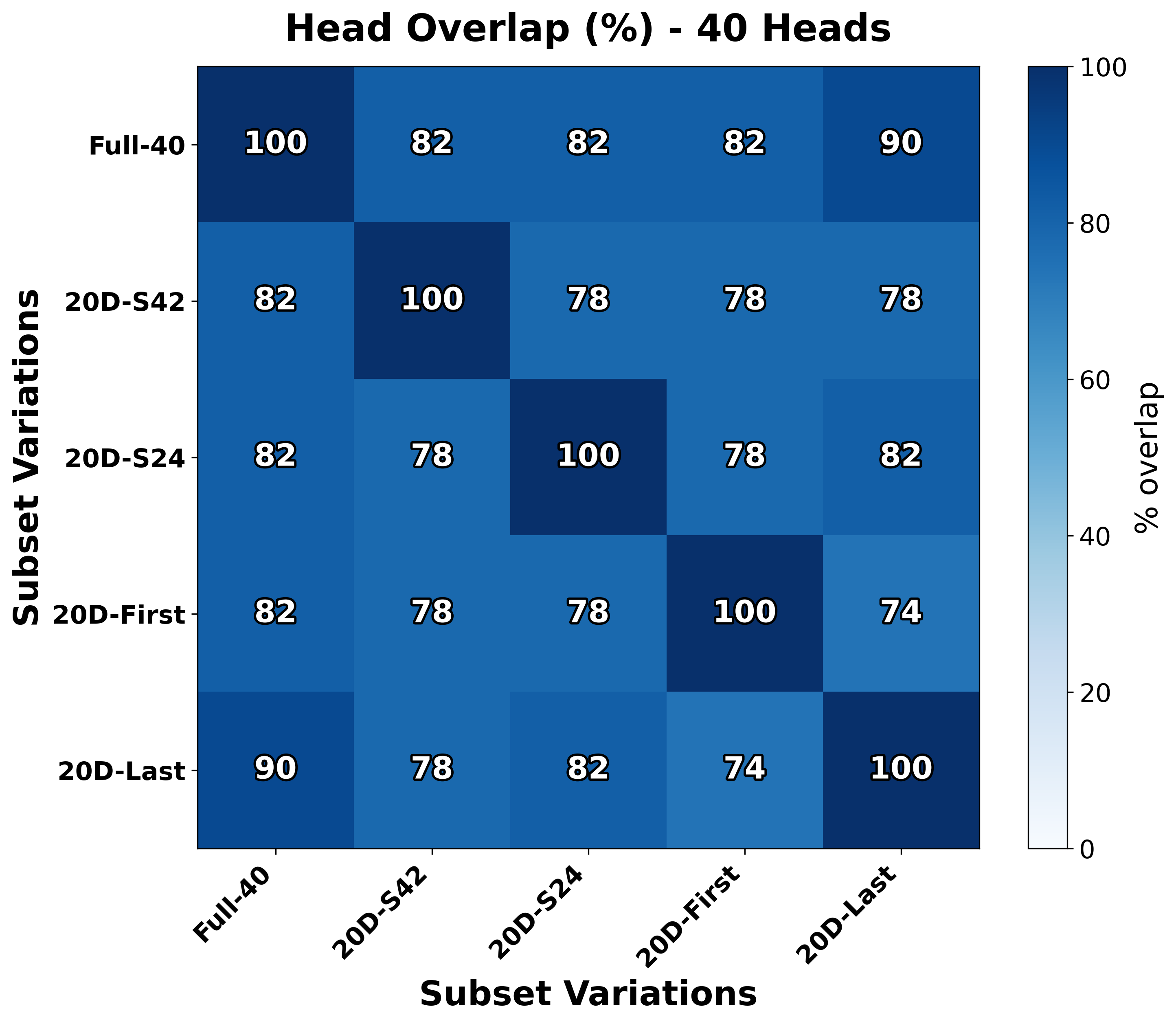}
  \caption{Top-40 Head Selection}
  \label{fig:head_overlap_40}
\end{subfigure}
\caption{\textbf{Head Selection Consistency Across Data Subsets.} Heatmaps showing the percentage of overlapping attention heads selected from 5 different data variants (Full-40: all 40 demos; 20D-S42/S24: 20 randomly sampled demos with different seeds; 20D-First/Last: first/last 20 demos.) for (a) top-20 heads and (b) top-40 heads. }
\label{fig:head_consistency}
\end{figure*}

Figure~\ref{fig:head_consistency} shows the overlap percentage between selected heads. For top-20 heads, we observe 54\%-82\% pairwise overlap, with the full 40-demo set showing highest consistency (60\%-82\%). For top-40 heads, the overlap increases to 74\%-90\%. This difference is expected: heads near the selection boundary (ranks 15-25) often have similar importance scores, so minor data variations can shift which specific heads fall within the top-20 cutoff. Top-40 contains higher overlap, confirming that our method reliably identifies the core task-relevant heads, with variations mainly in the borderline cases.

\minisection{Head Selection via Classification}
\label{supp:experiments:classification}
This method employs a binary classification approach to identify task-relevant attention heads. We have a small support set that contains 20 episodes for both positive (target task) and negative (non-target tasks). Then, for each head, we independently compute class centroids by averaging activations across the provided positive and negative samples.

Each head is scored based on its discriminative ability using margin-based metrics, which is computed as the difference between each head's similarity to the positive class centroid and its similarity to the negative class centroid for all support set samples (with signs flipped for negative samples). The top-k highest-scoring heads are selected as the sparse subset, typically reducing from 144 to k (here we pick 20) heads while having stronger task discrimination than all heads. During inference, selected heads perform majority voting where each head contributes a vote based on cosine similarity to learned centroids, effectively capturing task-specific semantic patterns with significantly reduced computational overhead compared to other comparable feature selection or sparsifcation methods.

As shown in Table~\ref{tab:SAV}, while the top 20 selected heads achieve the highest performance (86.75\% accuracy) compared to using all 144 heads (83.5\% accuracy), the improvement is modest. More importantly, random head selection yields surprisingly competitive performance (82.5\% accuracy), with only a 4.25\% gap compared to the top-performing heads. Even the worst 20 heads achieve reasonable accuracy (81.25\%), indicating that most attention heads possess some degree of task-relevant discriminative capability. 

The voting margin column reported in the performance evaluation table represents the difference between correct and incorrect votes across all selected heads for each sample, with values ranging from $-K$ to $K$ where $K$ denotes the number of heads we choose to select.

This observation suggests that the distinction between "best" and "worst" heads is not sufficiently pronounced for effective head selection. The relatively small performance gap across different head subsets implies that the task-relevant information is distributed across most attention heads rather than concentrated in a few specialized ones. Consequently, we did not adopt this classification-based head selection approach, as the limited discriminative ability between heads undermines the fundamental assumption that sparse head selection can significantly improve performance while reducing computational overhead.

\begin{table}[t]
\centering
\caption{Classification Performance of Selected Attention Heads}
\label{tab:SAV}
\begin{tabular}{lcc}
\toprule
\textbf{Head Selection} & \textbf{Accuracy} & \textbf{Voting Margin} \\
\midrule
Top 20 heads & 86.75\% & 9.13 / 20 \\
Worst 20 heads & 81.25\% & 6.57 / 20 \\
All 144 heads & 83.5\% & 45.62 / 144 \\
Random 20 heads & 82.5\% & 6.75 / 20 \\
\bottomrule
\end{tabular}
\end{table}

\begin{figure}[t]
\centering
\includegraphics[width=\columnwidth]{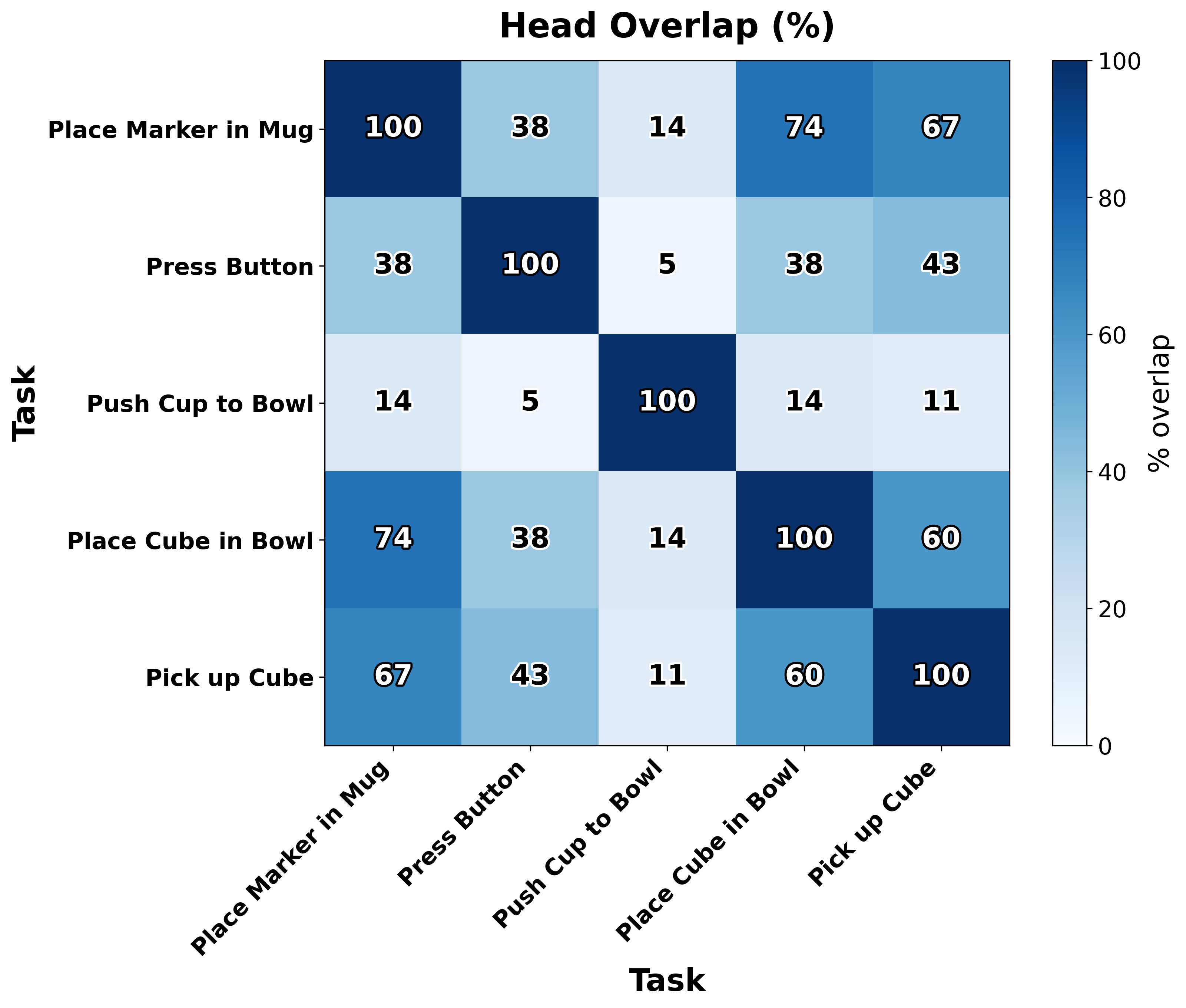}
\caption{\textbf{Cross-Task Head Selection Overlap.} Heatmap showing the percentage of overlapping attention heads selected across different tasks.}
\label{fig:head_overlap}
\end{figure}

\minisection{Head Selection Overlapping across Tasks} Figure~\ref{fig:head_overlap} shows that the selected heads have relatively low overlap across tasks (5--74\% similarity), demonstrating that our method identifies task-specific rather than generic attention heads. 
Moreover, the overlap is structured: tasks that are semantically similar share more heads, e.g., \textit{Place Marker in Mug} and \textit{Place Green Cube in Red Bowl} exhibit a relatively high overlap (74\%), whereas more dissimilar tasks show only limited overlap. This result provides some support for the idea that the sparse representations identified by our method are grounded to the semantics of the physical task.

\subsection{Additional Visualizations on Head Selection}
\label{supp:additional visualization}
\begin{figure*}[t]
    \centering
    \begin{subfigure}[b]{0.32\textwidth}
        \centering
        \includegraphics[width=\textwidth]{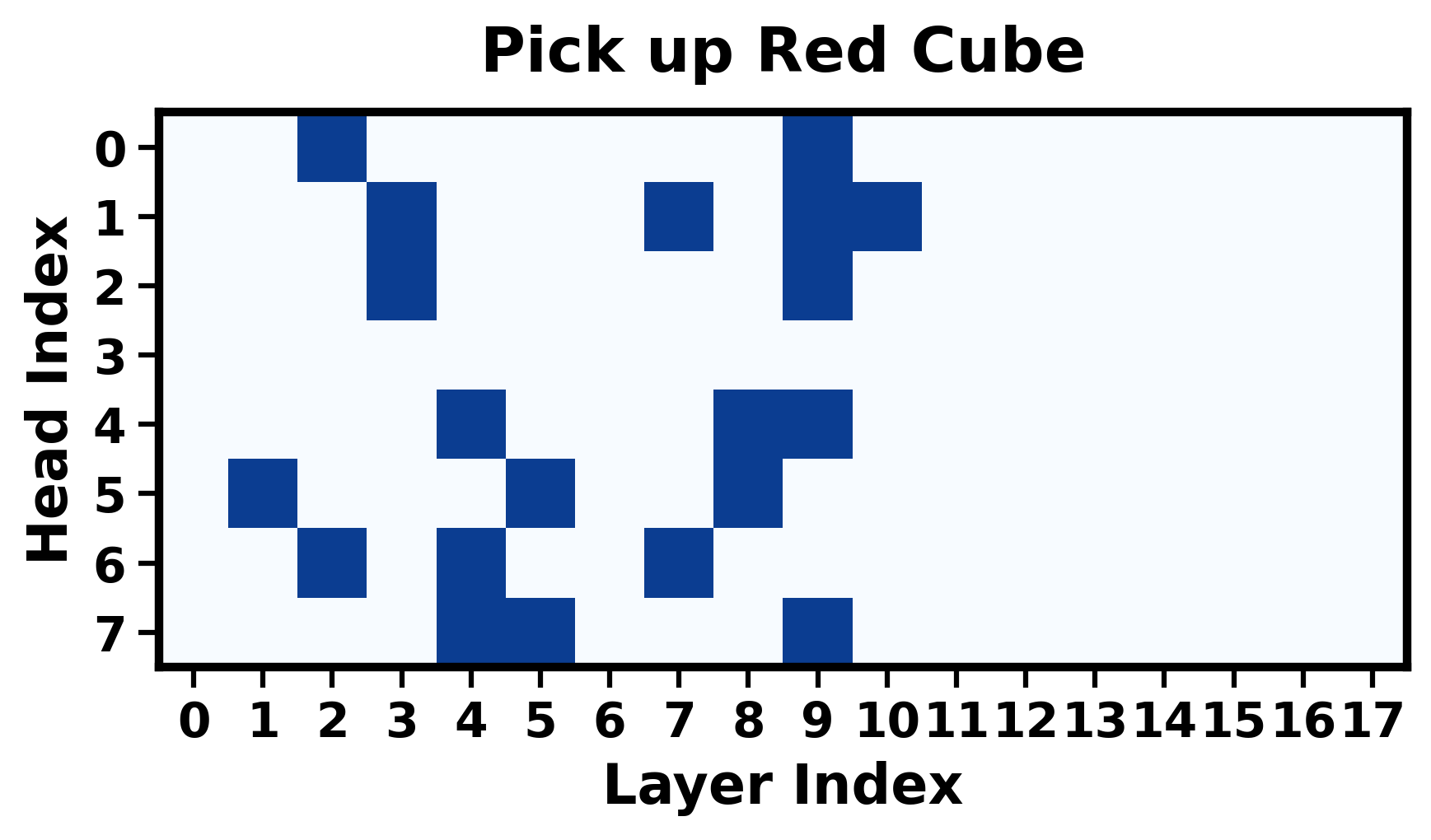}
        \caption{Pick Up Red Cube}
        \label{fig:sub:a_vertical}
    \end{subfigure}
    \hfill
    \begin{subfigure}[b]{0.32\textwidth}
        \centering
        \includegraphics[width=\textwidth]{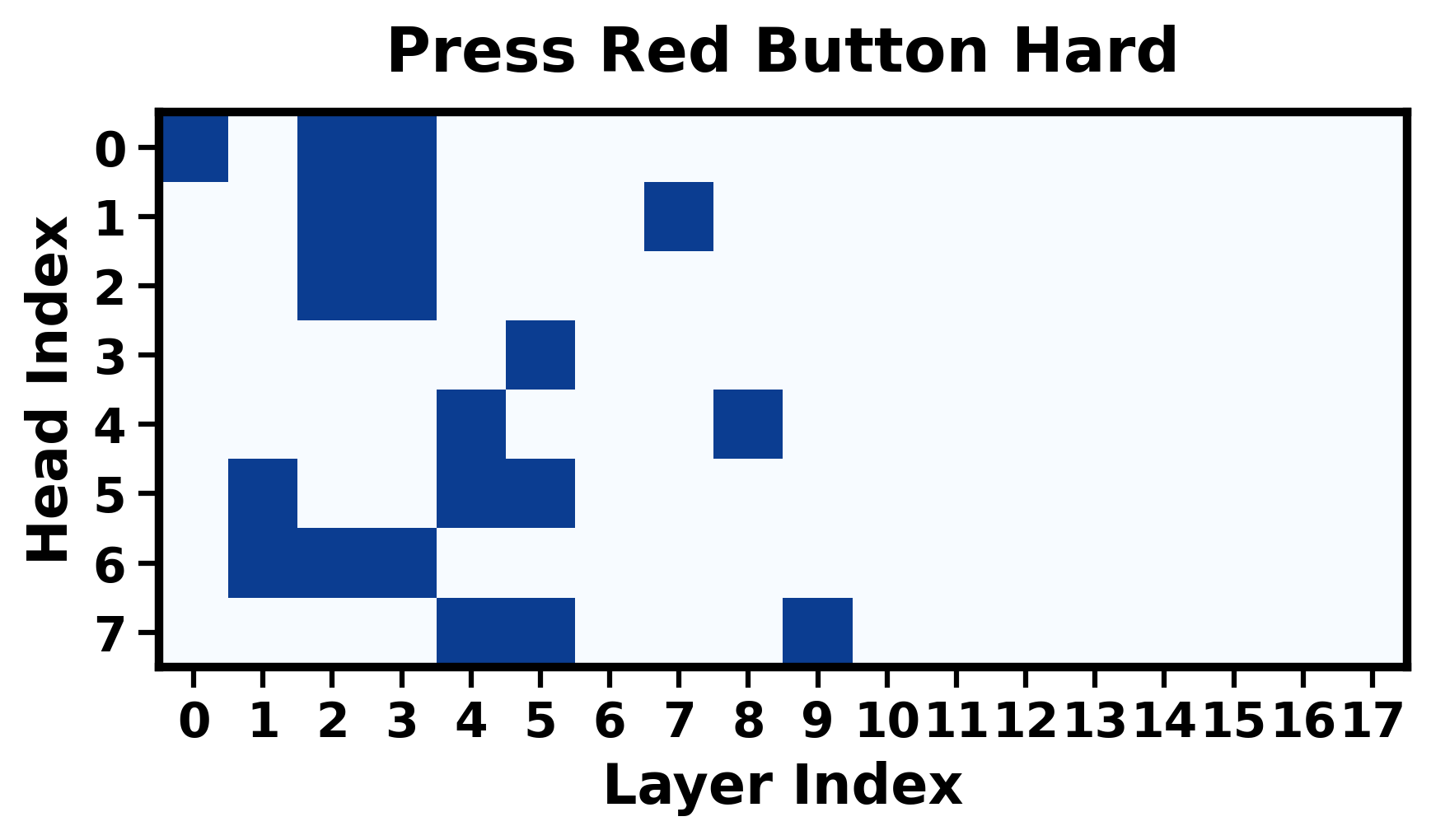}
        \caption{Press Red Button Hard}
        \label{fig:sub:b_vertical}
    \end{subfigure}
    \hfill
    \begin{subfigure}[b]{0.32\textwidth}
        \centering
        \includegraphics[width=\textwidth]{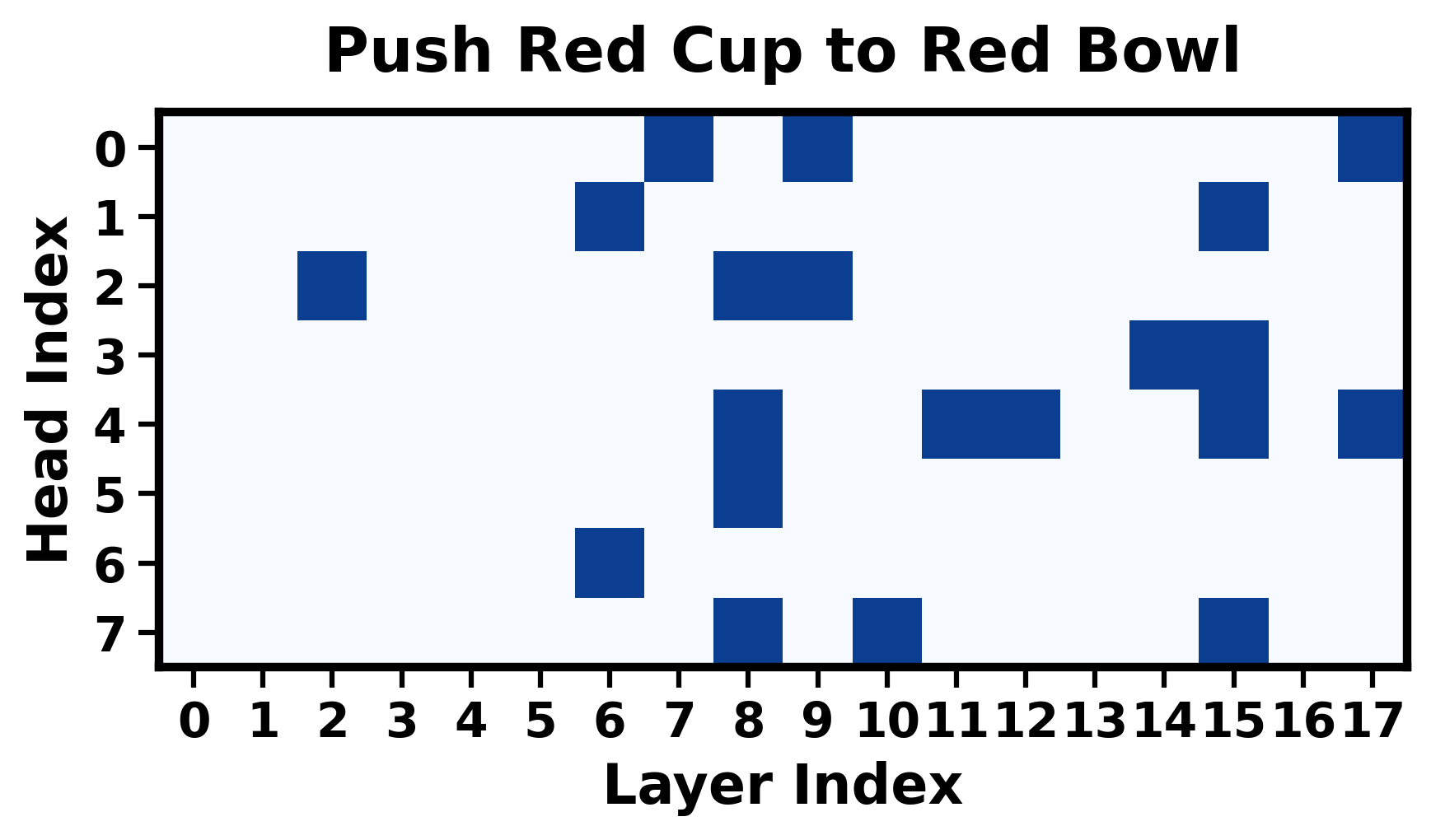}
        \caption{Push Red Cup to Red Bowl}
        \label{fig:sub:c_vertical}
    \end{subfigure}
    
    \caption{\textbf{Attention Head Selection on Additional Tasks.} Selected attention heads for (a) Pick Up Red Cube, (b) Press Red Button Hard, and (c) Push Red Cup to Red Bowl.}
    \label{fig:head_visual_more}
\end{figure*}

Figure~\ref{fig:head_visual_more} shows the heads selected in 3 additional tasks: Pick up Red Cube, Press Red Button Hard, and Push Red Cup to Red Bowl.

\section{Implementation Additional Details}
\label{supp:implementation}

\subsection{Fine-tuning Details}
\label{supp:fine-tuning setup}
In this setup we perform Robotic Steering with a mask-aware optimizer over LoRA adapters. All non-LoRA weights in the Gemma backbone are frozen. The SigLIP encoder and the entire Action Expert branch are also frozen. We update only the LoRA parameters at selected query and output attention heads, while freezing KV LoRA as they are shared per layer. The mask ensures gradient updates apply only to targeted head slices.

Beyond the selected heads, we allow a small set of auxiliary modules to update: action projections, time-conditioning MLPs, state projection ($\pi_0$ only), normalization adapters, and the main LLM feed-forward layers. As a baseline, we compare against standard LoRA fine-tuning following the official $\pi_0$/$\pi_0.5$ setup, where all LoRA adapters and the SigLIP encoder are trainable. Table~\ref{tab:trainable_modules} details the trainable components of both approaches.

\begin{table*}[t]
\centering
\small
\caption{Comparison of trainable modules between standard LoRA fine-tuning and our Robotic Steering approach. \trained~indicates trainable, \frozen~indicates frozen.}
\label{tab:trainable_modules}
\begin{tabularx}{\textwidth}{@{}lXccc@{}}
\toprule
\textbf{Category} & \textbf{Module} & \textbf{Simple LoRA} & \textbf{Robotic Steering ($\pi_0$)} & \textbf{Robotic Steering ($\pi_{0.5}$)} \\
\midrule
\multirow{2}{*}{\textbf{Main LLM -- Attention}} 
& Q \& O heads (LoRA) & All heads & Selected only & Selected only \\
& K \& V projections (LoRA) & \trained & \frozen & \frozen \\
\midrule
\multirow{2}{*}{\textbf{Main LLM -- FFN \& Norm}} 
& Feed-forward (\texttt{mlp}) & \trained & \trained & \trained \\
& Normalization adapters & \trained & \trained~(scale) & \trained~(Dense) \\
\midrule
\multirow{3}{*}{\textbf{Action Expert Branch}} 
& Attention (Q/K/V/O LoRA) & \trained & \frozen & \frozen \\
& Feed-forward (\texttt{mlp\_1}) & \trained & \frozen & \frozen \\
& Normalization (\texttt{*norm\_1}) & \trained & \frozen & \frozen \\
\midrule
\textbf{Vision Encoder} 
& SigLIP image encoder & \trained & \frozen & \frozen \\
\midrule
\multirow{3}{*}{\textbf{Diffusion Adapters}} 
& Action in/out projections & \trained & \trained & \trained \\
& Time conditioning MLP & \trained & \trained & \trained \\
& State projection & \trained & \trained & - \\
\bottomrule
\end{tabularx}
\end{table*}

\subsection{Observation and Action Space}
 We take both wrist image and external right image plus and 7 joint positions and 1 gripper position as observation input. All images are (1080,720), we first center-crop them to (720,720) and then resize to (224, 224). And for action space we are using 7 joint velocities and 1 gripper position as action space.
\subsection{Key Hyperparameters for Fine-tuning}
We perform Robotic Steering fine-tuning for 5000 timesteps with a CosineDecaySchedule as following:
\begin{itemize}
  \item \textbf{Warmup steps:} 200
  \item \textbf{Peak learning rate:} $2.5\times 10^{-5}$
  \item \textbf{Decay steps:} 5000
  \item \textbf{Final learning rate:} $2.5\times 10^{-6}$
  \item \textbf{Total training steps:} 5000
  \item \textbf{Batch size:} 32
\end{itemize}
\section{Robotic Task Setup Additional Details}
\label{supp:robot_setup}


\subsection{Place Marker in Cup}
\noindent\textbf{Task:} Grasp a small marker and place it inside a target cup.

\noindent\textbf{Success Criteria:} Marker is fully contained within the cup.

\subsection{Press Red Button Hard}
\noindent\textbf{Task:} Locate and press a red button with sufficient force to activate it.

\noindent\textbf{Success Criteria:} Button is pressed hard enough to trigger the mechanism.

\subsection{Place Green Cube in Red Bowl}
\noindent\textbf{Task:} Pick up a green cube and place it into a red bowl.

\noindent\textbf{Success Criteria:} Green cube is fully contained within the red bowl.

\subsection{Pick Up Red Cube}
\noindent\textbf{Task:} Grasp and lift a red cube from the surface.

\noindent\textbf{Success Criteria:} Red cube is stably grasped and lifted clear of the surface.

\subsection{Push Red Cup to Red Bowl}
\noindent\textbf{Task:} Push a red cup across the surface until it contacts a red bowl. The cup should remain upright and not fall over.
\noindent\textbf{Success Criteria:} Red cup makes contact with the red bowl.

\subsection{Experimental Variations}

\begin{figure}[t]
    \centering
    \includegraphics[width=\linewidth]{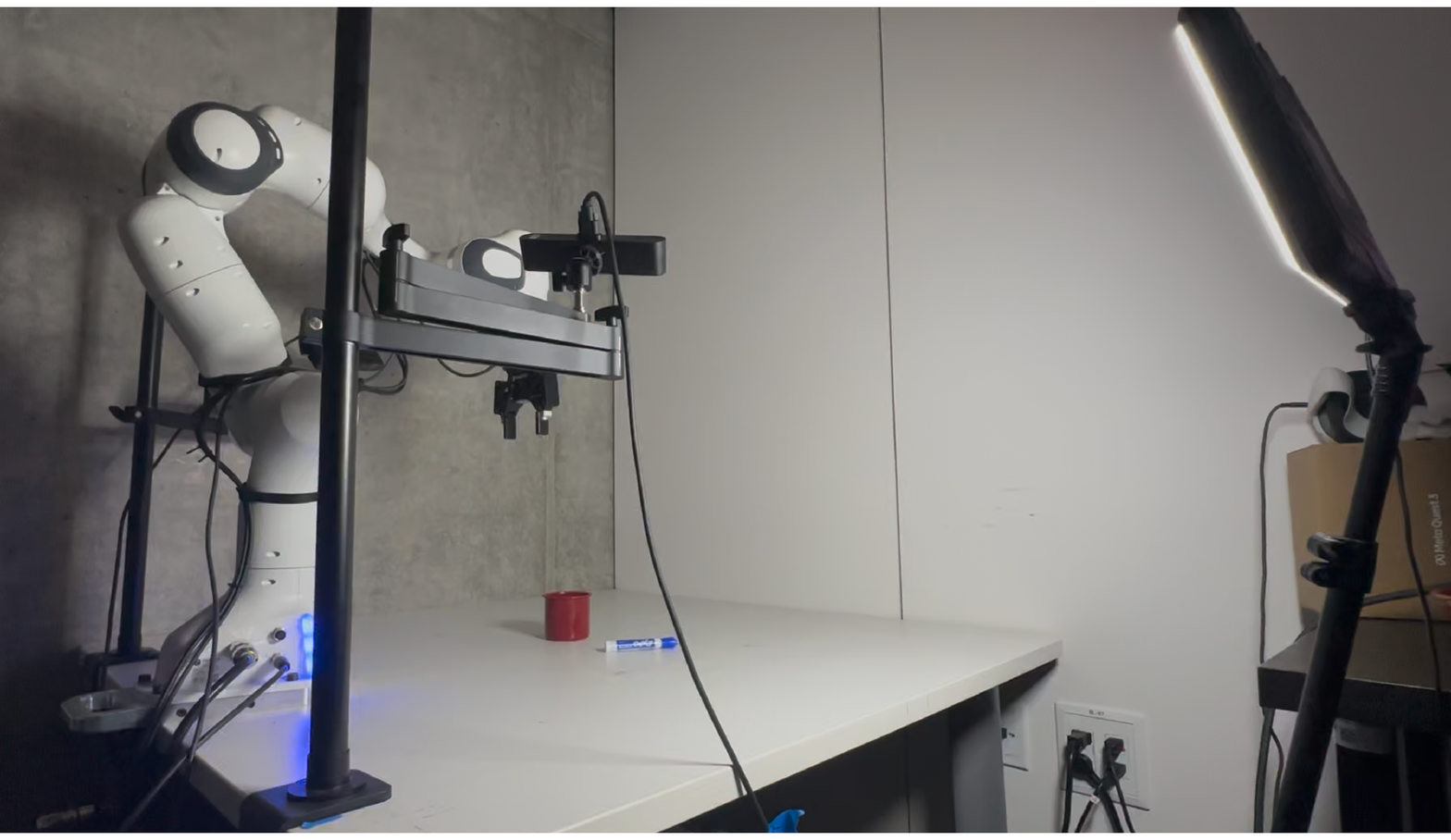}
    \caption{Lighting variation with dim ambient lighting and a bright directional lamp illuminating only one side of the workspace.}
    \label{fig:lighting_variation}
\end{figure}

\noindent\textbf{Lighting variation:} We dimmed the ceiling lights and introduced a bright, directional white lamp positioned to one side of the workspace. This created strong illumination on the side facing the lamp and pronounced shadows on the opposite side, emphasizing high-contrast edges and asymmetric shading. This setup tests how robust the model is to extreme lighting conditions and features that are only clearly visible from one side of the scene, as illustrated in Figure~\ref{fig:lighting_variation}.

\begin{figure}[t]
    \centering
    \includegraphics[width=0.8\columnwidth]{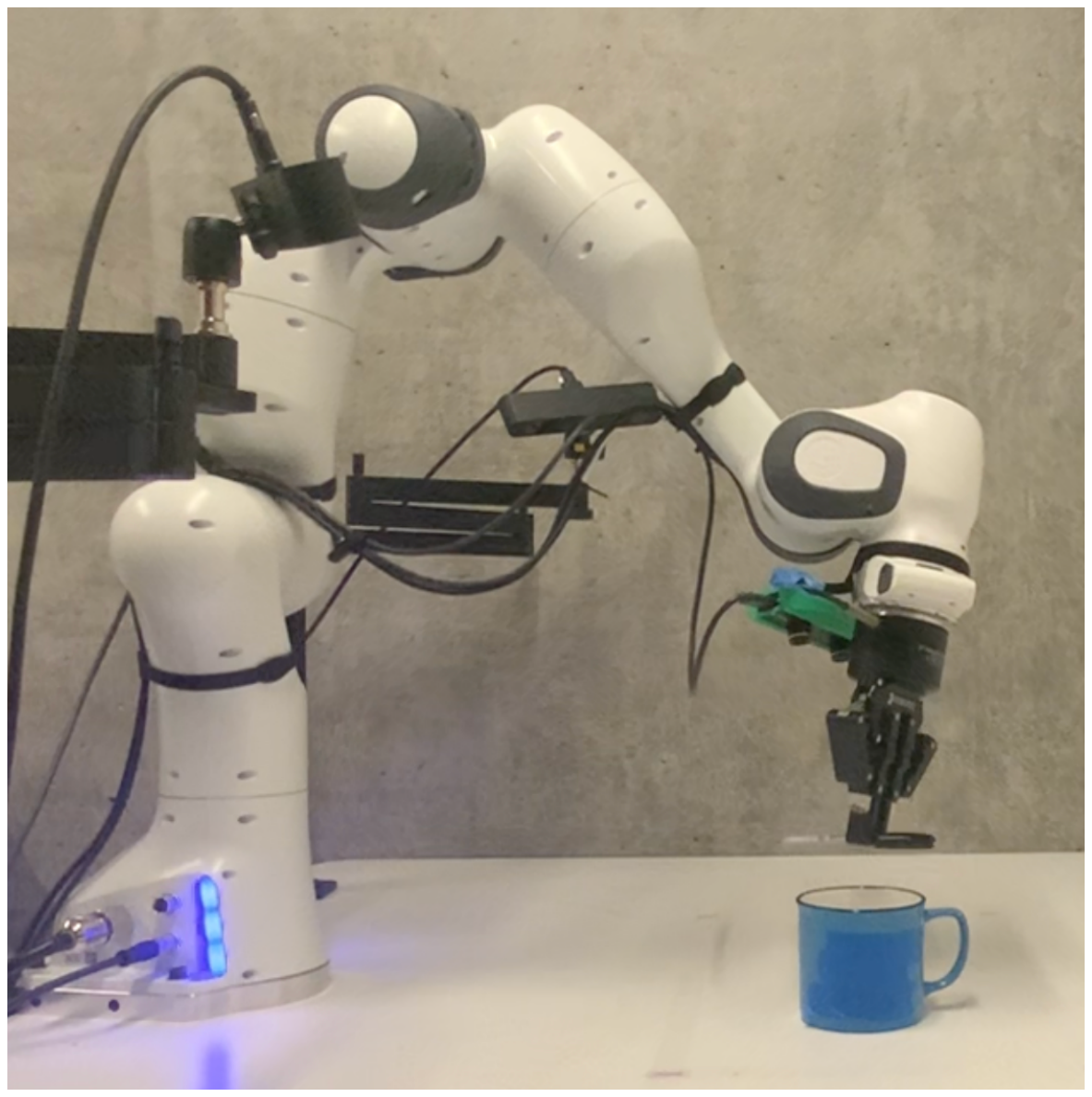}
    \caption{Form-change variation where the mug color and marker shape are modified to evaluate robustness to changes in object appearance.}
    \label{fig:form_change}
\end{figure}

\noindent\textbf{Form-change variation:} We altered the appearance of the manipulated objects by changing the color of the mug and modifying the marker's 3D geometry (e.g., length, thickness, and tip shape). These changes modify the visual and geometric cues available to the model, providing a stress test for how well it transfers to objects with different colors and silhouettes while preserving the underlying task semantics. Figure~\ref{fig:form_change} shows examples of these modifications.

\begin{figure}[ht]
    \centering
    \includegraphics[width=0.9\columnwidth]{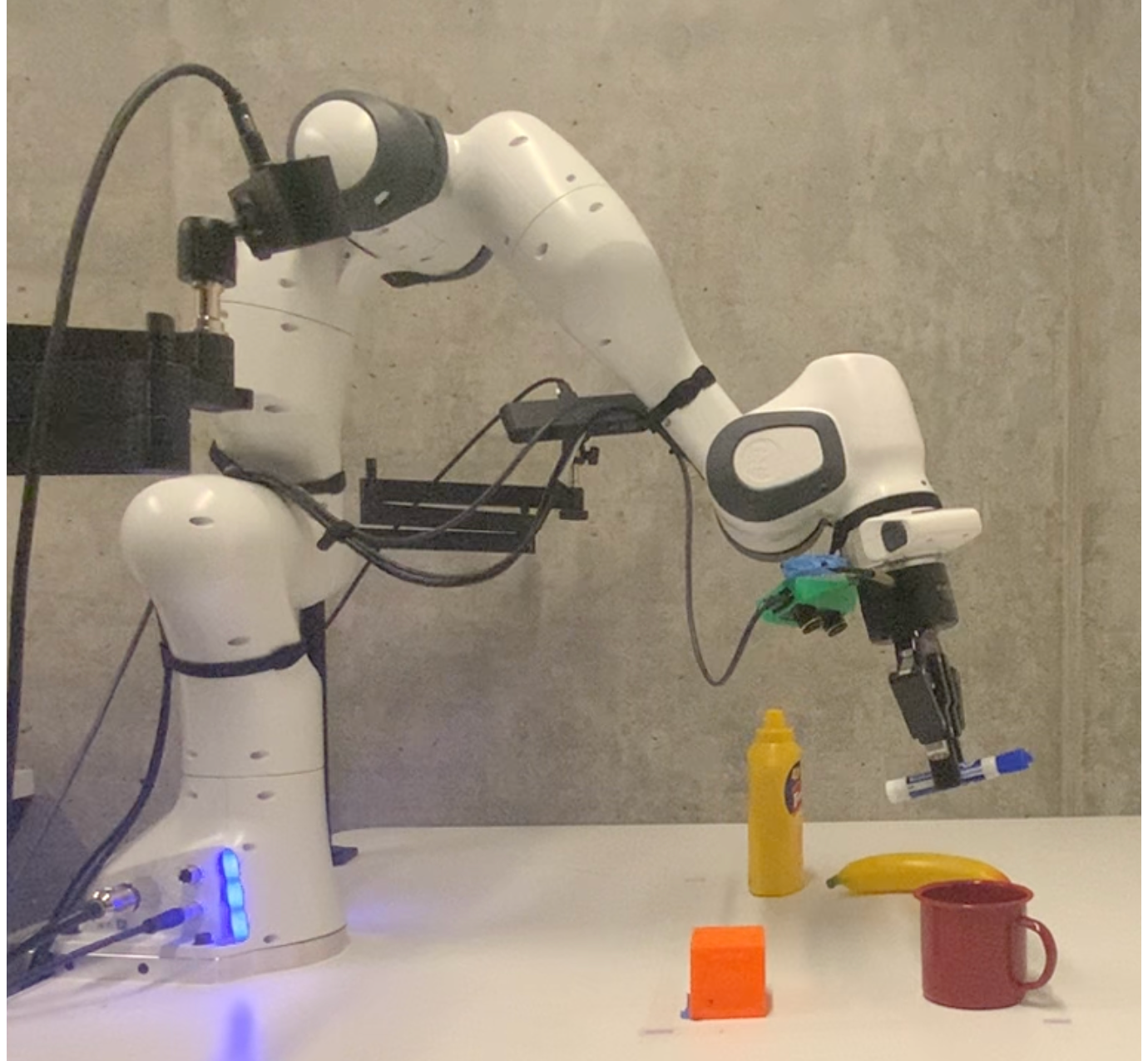}
    \caption{Distractor variation with additional objects placed around the workspace to introduce clutter and competing visual features.}
    \label{fig:distractor_variation}
\end{figure}

\noindent\textbf{Distractor variation:} We introduced additional objects into the workspace to act as visual distractors around the initial and final targets. These distractors add clutter, extra edges, and potential occlusions that can confuse the visual encoder. This variation evaluates whether the model can reliably focus on the correct targets amidst competing features and avoid being misled by irrelevant items in the scene. An example setup is shown in Figure~\ref{fig:distractor_variation}.

\section{Franka Robot Setup and Data Collection}
\label{supp:franka_setup}

\subsection{Robot Hardware}

\begin{figure}[ht]
    \centering
    \includegraphics[width=0.9\columnwidth]{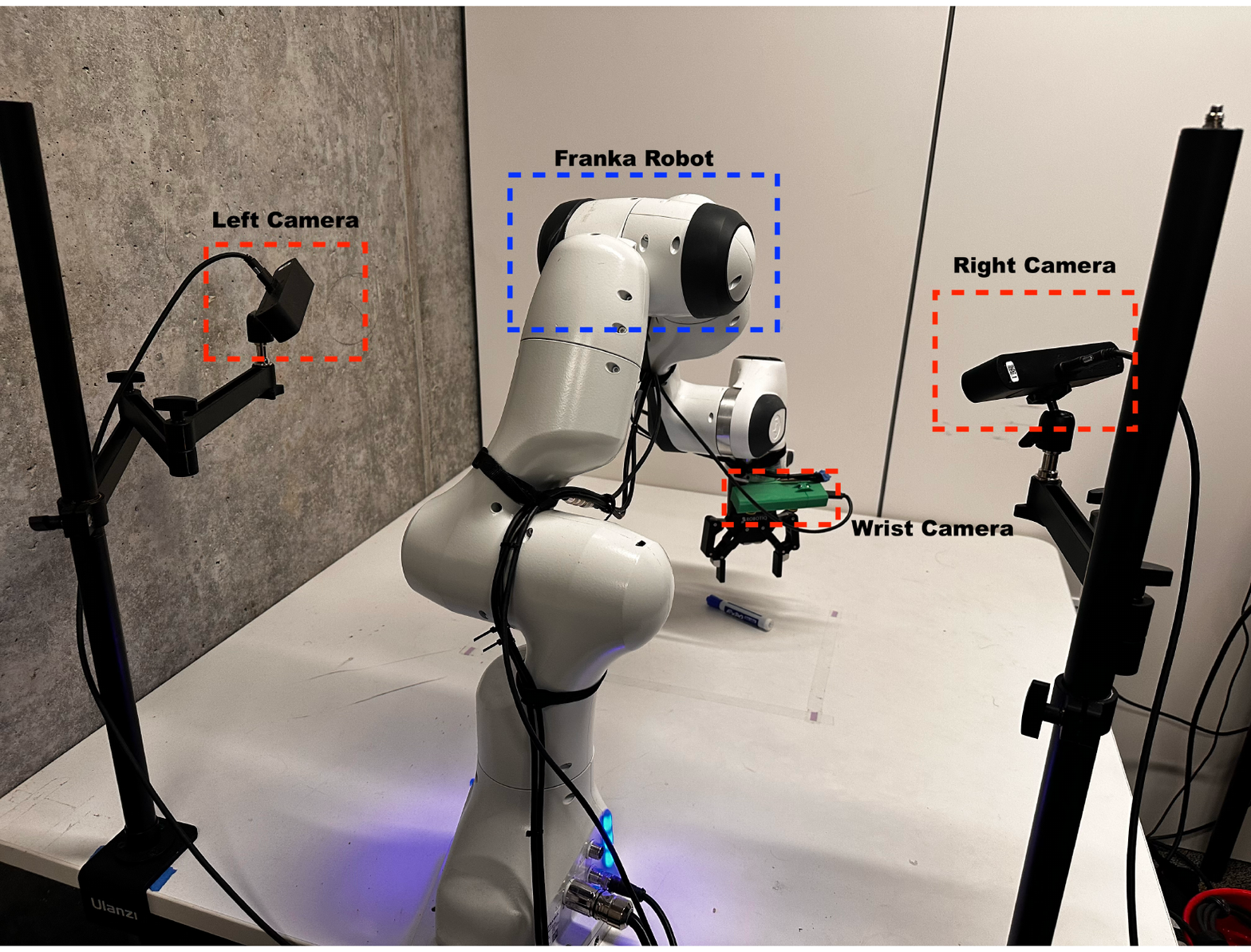} 
    \caption{Hardware configuration for the Franka Emika Panda robot with Robotiq gripper, wrist camera, and left/right side cameras used during data collection.}
    \label{fig:supp_frank_setup}
\end{figure}

Our experiments use a Franka Emika Panda arm with seven actuated joints and a RobotiQ parallel gripper at the end-effector. A ZED~2i camera is mounted on the wrist to provide an egocentric view of the scene during interaction. Two additional ZED~2i cameras are fixed to the left and right of the robot, giving wide-angle side views of the workspace. All cameras stream RGB video at 1280\,$\times$\,720 resolution and 60\,Hz without depth sensing. This multi-view configuration captures both close-up hand motion and the global scene layout used by our policy.The complete hardware setup is shown in Figure~\ref{fig:supp_frank_setup}.

\subsection{Teleoperation Interface}
We collect demonstrations by teleoperating the arm with a Meta Quest~3 headset. Only the right-hand controller is mapped to the robot, which controls the end-effector pose and gripper through a smooth Cartesian impedance controller. Demonstrations are executed in a single continuous motion such that accidental impacts do not damage the objects or interrupt the trajectory. The low-latency VR interface makes it easy for the operator to perform precise pick-and-place and pushing behaviors.

\subsection{Data Collection and Evaluation Protocol}
During data collection we record synchronized RGB video from the wrist and both side cameras, along with the robot's joint positions, velocities, and gripper state. We adopt the Franka configuration and control limits used in the Franka-DROID setup of Khazatsky et al.\ (2024) to ensure safe operation.

For quantitative evaluation, we define a rectangular region of size $0.5 \times 0.28$\,m on the table in front of the robot and discretize it into a $5 \times 8$ grid, yielding 40 distinct object placements. For each trial, we place the object at one grid location and randomize its yaw angle about the vertical axis. A policy is evaluated by running rollouts for every grid cell and computing the success rate over all 40 placements for each object, and then averaging these scores across objects to obtain the final performance measure.






















%% file: main.bib
@String(ECCV= {Eur. Conf. Comput. Vis.})

@String(NIPS= {Adv. Neural Inform. Process. Syst.})

@String(ICLR = {Int. Conf. Learn. Represent.})

@String(ECCV  = {ECCV})

@String(NIPS  = {NeurIPS})

@String(ICLR  = {ICLR})

@article{Niu2024LLARVAVI,
  title={LLARVA: Vision-Action Instruction Tuning Enhances Robot Learning},
  author={Dantong Niu and Yuvan Sharma and Giscard Biamby and Jerome Quenum and Yutong Bai and Baifeng Shi and Trevor Darrell and Roei Herzig},
  journal={ArXiv},
  year={2024},
  volume={abs/2406.11815},
  url={https://api.semanticscholar.org/CorpusID:270559839}
}

@inproceedings{Anthropic2023Claude,
  title={The Claude 3 Model Family: Opus, Sonnet, Haiku},
  author={},
  url={https://api.semanticscholar.org/CorpusID:268232499}
}

@article{hendel2023context,
  title={In-context learning creates task vectors},
  author={Hendel, Roee and Geva, Mor and Globerson, Amir},
  journal={arXiv preprint arXiv:2310.15916},
  year={2023}
}

@article{Jiang2023Mistral7,
  title={Mistral 7B},
  author={Albert Qiaochu Jiang and Alexandre Sablayrolles and Arthur Mensch and Chris Bamford and Devendra Singh Chaplot and Diego de Las Casas and Florian Bressand and Gianna Lengyel and Guillaume Lample and Lucile Saulnier and L'elio Renard Lavaud and Marie-Anne Lachaux and Pierre Stock and Teven Le Scao and Thibaut Lavril and Thomas Wang and Timoth{\'e}e Lacroix and William El Sayed},
  journal={ArXiv},
  year={2023},
  volume={abs/2310.06825},
}

@article{bai2023qwen,
  title={Qwen technical report},
  author={Bai, Jinze and Bai, Shuai and Chu, Yunfei and Cui, Zeyu and Dang, Kai and Deng, Xiaodong and Fan, Yang and Ge, Wenbin and Han, Yu and Huang, Fei and others},
  journal={arXiv preprint arXiv:2309.16609},
  year={2023}
}

@inproceedings{huang2024multimodal,
 title = {Multimodal Task Vectors Enable Many-Shot Multimodal In-Context Learning},
author = {Huang, Brandon and Mitra, Chancharik and Arbelle, Assaf and Karlinsky, Leonid and Darrell, Trevor and Herzig, Roei},
 booktitle = {Advances in Neural Information Processing Systems},
 pages = {22124--22153},
 volume = {37},
 year = {2024}
}

@article{Touvron2023LLaMAOA,
  title={LLaMA: Open and Efficient Foundation Language Models},
  author={Hugo Touvron and Thibaut Lavril and Gautier Izacard and Xavier Martinet and Marie-Anne Lachaux and Timoth{\'e}e Lacroix and Baptiste Rozi{\`e}re and Naman Goyal and Eric Hambro and Faisal Azhar and Aurelien Rodriguez and Armand Joulin and Edouard Grave and Guillaume Lample},
  journal={ArXiv},
  year={2023},
  volume={abs/2302.13971},
}

@article{lora,
  title={Lora: Low-rank adaptation of large language models},
  author={Hu, Edward J and Shen, Yelong and Wallis, Phillip and Allen-Zhu, Zeyuan and Li, Yuanzhi and Wang, Shean and Wang, Lu and Chen, Weizhu},
  journal={arXiv preprint arXiv:2106.09685},
  year={2021}
}

@misc{liu2023llava15,
      title={Improved Baselines with Visual Instruction Tuning}, 
      author={Haotian Liu and Chunyuan Li and Yuheng Li and Yong Jae Lee},
      year={2023},
      eprint={2310.03744},
      archivePrefix={arXiv},
      primaryClass={cs.CV}
}

@article{reed2022generalist,
  title={A Generalist Agent},
  author={Reed, Scott and Zolna, Konrad and Parisotto, Emilio and Colmenarejo, Sergio Gomez and Novikov, Alexander and Barth-Maron, Gabriel and Gimenez, Mai and Sulsky, Yury and Kay, Jackie and Springenberg, Jost Tobias and others},
  journal={arXiv preprint arXiv:2205.06175},
  year={2022}
}

@string{iclr = " International Conference on Learning Representations (ICLR) "}

@string{eccv = "European Conference on Computer Vision (ECCV)"}

@string{emnlp = "Conference on Empirical Methods in Natural Language Processing (EMNLP)"}

@string{nips = "Advances in Neural Information Processing Systems (NeurIPS)"}

@inproceedings{liu2023llava,
    author      = {Liu, Haotian and Li, Chunyuan and Wu, Qingyang and Lee, Yong Jae},
    title       = {Visual Instruction Tuning},
    booktitle   = nips,
    year        = {2023}
}

@string{iclr = "International Conference on Learning Representations (ICLR)"}

@article{OpenAI2023GPT4TR,
  title={GPT-4 Technical Report},
  author={OpenAI},
  journal={ArXiv},
  year={2023},
  volume={abs/2303.08774},
}

@article{Bai2023QwenVLAF,
  title={Qwen-VL: A Frontier Large Vision-Language Model with Versatile Abilities},
  author={Jinze Bai and Shuai Bai and Shusheng Yang and Shijie Wang and Sinan Tan and Peng Wang and Junyang Lin and Chang Zhou and Jingren Zhou},
  journal={ArXiv},
  year={2023},
  volume={abs/2308.12966},
}

@article{team2023gemini,
  title={Gemini: a family of highly capable multimodal models},
  author={Team, Gemini and Anil, Rohan and Borgeaud, Sebastian and Wu, Yonghui and Alayrac, Jean-Baptiste and Yu, Jiahui and Soricut, Radu and Schalkwyk, Johan and Dai, Andrew M and Hauth, Anja and others},
  journal={arXiv preprint arXiv:2312.11805},
  year={2023}
}

@article{Reid2024Gemini1.5,
  title={Gemini 1.5: Unlocking multimodal understanding across millions of tokens of context},
  author={Team, Gemini},
  journal={ArXiv},
  year={2024},
  volume={abs/2403.05530},
}

@inproceedings{Hojel2024FindingVT,
  title={Finding visual task vectors},
  author={Hojel, Alberto and Bai, Yutong and Darrell, Trevor and Globerson, Amir and Bar, Amir},
  booktitle={European Conference on Computer Vision (ECCV)},
  pages={257--273},
  year={2025},
  organization={Springer}
}

@article{olsson2022context,
  title={In-context learning and induction heads},
  author={Olsson, Catherine and Elhage, Nelson and Nanda, Neel and Joseph, Nicholas and DasSarma, Nova and Henighan, Tom and Mann, Ben and Askell, Amanda and Bai, Yuntao and Chen, Anna and others},
  journal={arXiv preprint arXiv:2209.11895},
  year={2022}
}

@article{kanwisher2000funcspec,
  title={Domain specificity in face perception},
  author={Kanwisher, Nancy},
  journal={Nature neuroscience},
  volume={3},
  number={8},
  pages={759--763},
  year={2000},
  publisher={Nature Publishing Group}
}

@article{
FedorenkoFuncSpec,
author = {Evelina Fedorenko  and Michael K. Behr  and Nancy Kanwisher },
title = {Functional specificity for high-level linguistic processing in the human brain},
journal = {Proceedings of the National Academy of Sciences},
volume = {108},
number = {39},
pages = {16428-16433},
year = {2011},
doi = {10.1073/pnas.1112937108}}

@inproceedings{oggpt,
  title={Improving Language Understanding by Generative Pre-Training},
  author={Alec Radford and Karthik Narasimhan},
  year={2018},
  url={https://api.semanticscholar.org/CorpusID:49313245}
}

@inproceedings{subramani-etal-2022-extracting,
  title = {Extracting Latent Steering Vectors from Pretrained Language Models},
  author = {Subramani, Nishant and Suresh, Nivedita and Peters, Matthew},
  booktitle = {Findings of the Association for Computational Linguistics: ACL 2022},
  month = may,
  year = {2022},
  address = {Dublin, Ireland},
  publisher = {Association for Computational Linguistics},
  pages = {566--581}
}

@inproceedings{hendel-etal-2023-context,
  title = {In-Context Learning Creates Task Vectors},
  author = {Hendel, Roee and Geva, Mor and Globerson, Amir},
  booktitle = {Findings of the Association for Computational Linguistics: EMNLP 2023},
  month = dec,
  year = {2023},
  address = {Singapore},
  publisher = {Association for Computational Linguistics},
  pages = {9318--9333}
}

@inproceedings{Todd2024function,
  title = {Function Vectors in Large Language Models},
  author = {Todd, Eric and Li, Millicent L. and Sharma, Arnab Sen and Mueller, Aaron and Wallace, Byron C. and Bau, David},
  booktitle = {Proceedings of the International Conference on Learning Representations (ICLR)},
  year = {2024}
}

@inproceedings{Hojel2024visual,
  title = {Finding Visual Task Vectors},
  author = {Hojel, Alberto and Bai, Yutong and Darrell, Trevor and Globerson, Amir and Bar, Amir},
  booktitle = {Proceedings of the European Conference on Computer Vision (ECCV)},
  year = {2024}
}

@article{Hernandez2023Inspecting,
  title   = {Inspecting and Editing Knowledge Representations in Language Models},
  author  = {Evan Hernandez and Belinda Z. Li and Jacob Andreas},
  journal = {arXiv preprint arXiv:2304.00740},
  year    = {2023},
  url     = {https://arxiv.org/abs/2304.00740}
}

@article{Turner2024ActivationSteering,
  title   = {Steering Language Models with Activation Engineering},
  author  = {Alexander Matt Turner and Lisa Thiergart and Gavin Leech and David Udell and Juan J. Vazquez and Ulisse Mini and Monte MacDiarmid},
  journal = {arXiv preprint arXiv:2308.10248},
  year    = {2024},
  url     = {https://arxiv.org/abs/2308.10248}
}

@article{Panickssery2023CAA,
  title   = {Steering Llama 2 via Contrastive Activation Addition},
  author  = {Nina Panickssery and Nick Gabrieli and Julian Schulz and Meg Tong and Evan Hubinger and Alexander Matt Turner},
  journal = {arXiv preprint arXiv:2312.06681},
  year    = {2023},
  url     = {https://arxiv.org/abs/2312.06681}
}

@article{mitra2024sparse,
  title={Sparse Attention Vectors: Generative Multimodal Model Features Are Discriminative Vision-Language Classifiers},
  author={Mitra, Chancharik and Huang, Brandon and Chai, Tianning and Lin, Zhiqiu and Arbelle, Assaf and Feris, Rogerio and Karlinsky, Leonid and Darrell, Trevor and Ramanan, Deva and Herzig, Roei},
  journal={arXiv preprint arXiv:2412.00142},
  year={2024}
}

@inproceedings{bau2017network,
  title={Network Dissection: Quantifying Interpretability of Deep Visual Representations},
  author={Bau, David and Zhou, Bolei and Khosla, Aditya and Oliva, Aude and Torralba, Antonio},
  booktitle={Proceedings of the IEEE Conference on Computer Vision and Pattern Recognition},
  pages={6541--6549},
  year={2017}
}

@inproceedings{bau2020understanding,
  title={Understanding the Role of Individual Units in a Deep Neural Network},
  author={Bau, David and Zhou, Bolei and Khosla, Aditya and Oliva, Aude and Torralba, Antonio},
  booktitle={Proceedings of the National Academy of Sciences},
  volume={117},
  number={48},
  pages={30071--30077},
  year={2020}
}

@inproceedings{zhou2018interpreting,
  title={Interpreting Deep Visual Representations via Network Dissection},
  author={Zhou, Bolei and Bau, David and Oliva, Aude and Torralba, Antonio},
  booktitle={IEEE Transactions on Pattern Analysis and Machine Intelligence},
  volume={41},
  number={9},
  pages={2131--2145},
  year={2018}
}

@inproceedings{finn2017oneshot,

  title={One-Shot Visual Imitation Learning via Meta-Learning},

  author={Finn, Chelsea and Yu, Tianhe and Zhang, Tianhao and Abbeel, Pieter and Levine, Sergey},

  booktitle={Conference on Robot Learning (CoRL)},

  year={2017}

}

@inproceedings{yu2018one,

  title={One-Shot Imitation from Observing Humans via Domain-Adaptive Meta-Learning},

  author={Yu, Tianhe and Finn, Chelsea and Dasari, Sudeep and Xie, Annie and Zhang, Tianhao and Abbeel, Pieter and Levine, Sergey},

  booktitle={Robotics: Science and Systems (RSS)},

  year={2018}

}

@inproceedings{inproceedings,

author = {Ghadirzadeh, Ali and Chen, Xi and Poklukar, Petra and Finn, Chelsea and Björkman, Mårten and Kragic, Danica},

year = {2021},

month = {09},

pages = {1274-1280},

title = {Bayesian Meta-Learning for Few-Shot Policy Adaptation Across Robotic Platforms},

doi = {10.1109/IROS51168.2021.9636628}

}

@inproceedings{Grover2025EnhancingGI, title={Enhancing Generalization in Vision-Language-Action Models by Preserving Pretrained Representations}, author={Shresth Grover and Akshay Gopalkrishnan and Bo Ai and Henrik I. Christensen and Hao Su and Xuanlin Li}, year={2025}, url={https://api.semanticscholar.org/CorpusID:281315107} }

@article{hu2024prompt,
  title={Prompt tuning with diffusion for few-shot pre-trained policy generalization},
  author={Hu, Shengchao and Zhao, Wanru and Lin, Weixiong and Shen, Li and Zhang, Ya and Tao, Dacheng},
  journal={arXiv preprint arXiv:2411.01168},
  year={2024}
}

@inproceedings{yin2025context,
  title={In-context learning enables robot action prediction in llms},
  author={Yin, Yida and Wang, Zekai and Sharma, Yuvan and Niu, Dantong and Darrell, Trevor and Herzig, Roei},
  booktitle={2025 IEEE International Conference on Robotics and Automation (ICRA)},
  pages={8972--8979},
  year={2025},
  organization={IEEE}
}

@article{song2025few,
  title={Few-Shot Vision-Language Action-Incremental Policy Learning},
  author={Song, Mingchen and Deng, Xiang and Zhong, Guoqiang and Lv, Qi and Wan, Jia and Li, Yinchuan and Hao, Jianye and Guan, Weili},
  journal={arXiv preprint arXiv:2504.15517},
  year={2025}
}

@article{du2023behavior,
  title={Behavior retrieval: Few-shot imitation learning by querying unlabeled datasets},
  author={Du, Maximilian and Nair, Suraj and Sadigh, Dorsa and Finn, Chelsea},
  journal={arXiv preprint arXiv:2304.08742},
  year={2023}
}

@article{li2025controlvla,
  title        = {ControlVLA: Few-shot Object-centric Adaptation for Pre-trained Vision-Language-Action Models},
  author       = {Puhao Li and Yingying Wu and Ziheng Xi and Wanlin Li and Yuzhe Huang and Zhiyuan Zhang and Yinghan Chen and Jianan Wang and Song-Chun Zhu and Tengyu Liu and Siyuan Huang},
  journal      = {arXiv preprint arXiv:2506.16211},
  year         = {2025},
  url          = {https://arxiv.org/abs/2506.16211}
}

@article{sridhar2025ricl,
  title        = {RICL: Adding In-Context Adaptability to Pre-Trained Vision-Language-Action Models},
  author       = {Kaustubh Sridhar and Souradeep Dutta and Dinesh Jayaraman and Insup Lee},
  journal      = {arXiv preprint arXiv:2508.02062},
  year         = {2025},
  url          = {https://arxiv.org/abs/2508.02062}
}

@article{ma2024gvl,
  title        = {Vision Language Models are In-Context Value Learners},
  author       = {Yecheng Jason Ma and Joey Hejna and Ayzaan Wahid and Chuyuan Fu and Dhruv Shah and Jacky Liang and Zhuo Xu and Sean Kirmani and Peng Xu and Danny Driess and Ted Xiao and Jonathan Tompson and Osbert Bastani and Dinesh Jayaraman and Wenhao Yu and Tingnan Zhang and Dorsa Sadigh and Fei Xia},
  journal      = {arXiv preprint arXiv:2411.04549},
  year         = {2024},
  doi          = {10.48550/arXiv.2411.04549},
  url          = {https://arxiv.org/abs/2411.04549}
}

@article{dasari2019robonet,
  title={RoboNet: Large-Scale Multi-Robot Learning},
  author={Sudeep Dasari and Frederik Ebert and Stephen Tian and Suraj Nair and Bernadette Bucher and Karl Schmeckpeper and Siddharth Singh and Sergey Levine and Chelsea Finn},
  journal={CoRR},
  volume={abs/1910.11215},
  year={2019}
}

@inproceedings{openxembodiment2024rtx,
  title={Open X-Embodiment: Robotic Learning Datasets and RT-X Models},
  author={Open-X Embodiment Collaboration},
  booktitle={Proceedings of the IEEE International Conference on Robotics and Automation (ICRA)},
  pages={6892--6903},
  year={2024},
  doi={10.1109/ICRA57147.2024.10611477}
}

@article{khazatsky2024droid,
  title={DROID: A Large-Scale In-The-Wild Robot Manipulation Dataset},
  author={Alexander Khazatsky and Karl Pertsch and Suraj Nair and Ashwin Balakrishna and Sudeep Dasari and Siddharth Karamcheti and ... and Sergey Levine and Chelsea Finn},
  journal={arXiv preprint arXiv:2403.12945},
  year={2024}
}

@inproceedings{zitkovich2023rt2,
  title={RT-2: Vision-Language-Action Models Transfer Web Knowledge to Robotic Control},
  author={Brianna Zitkovich and Tianhe Yu and Sichun Xu and Peng Xu and Ted Xiao and ... and Montserrat Gonzalez Arenas and Kehang Han},
  booktitle={Proc. of the 7th Conference on Robot Learning (CoRL)},
  pages={2165--2183},
  year={2023},
  publisher={PMLR}
}

@article{kim2024openvla,
  title={OpenVLA: An Open-Source Vision-Language-Action Model},
  author={Moo Jin Kim and Karl Pertsch and Siddharth Karamcheti and Ted Xiao and Ashwin Balakrishna and Suraj Nair and Rafael Rafailov and Ethan Foster and Grace Lam and Pannag Sanketi and Quan Vuong and Thomas Kollar and Benjamin Burchfiel and Russ Tedrake and Dorsa Sadigh and Sergey Levine and Percy Liang and Chelsea Finn},
  journal={arXiv preprint arXiv:2406.09246},
  year={2024}
}

@article{black2024pi0,
  title={$\\pi_{0}$: A Vision-Language-Action Flow Model for General Robot Control},
  author={Kevin Black and Noah Brown and Danny Driess and Adnan Esmail and Michael Equi and Chelsea Finn and Niccolo Fusai and Lachy Groom and Karol Hausman and Brian Ichter and Szymon Jakubczak and Tim Jones and Liyiming Ke and Sergey Levine and Adrian Li-Bell and Mohith Mothukuri and Suraj Nair and Karl Pertsch and Lucy Xiaoyang Shi and James Tanner and Quan Vuong and Anna Walling and Haohuan Wang and Ury Zhilinsky},
  journal={CoRR},
  volume={abs/2410.24164},
  year={2024}
}

@article{physicalintelligence2025pi0.5,
  title={$\\pi_{0.5}$: A Vision-Language-Action Model with Open-World Generalization},
  author={{Physical Intelligence} and Kevin Black and Noah Brown and Danny Driess and Chelsea Finn and Sergey Levine and others},
  journal={CoRR},
  volume={abs/2504.16054},
  year={2025}
}

@article{driess2023palme,
  title={PaLM-E: An Embodied Multimodal Language Model},
  author={Danny Driess and Fei Xia and Mehdi S.~M. Sajjadi and Corey Lynch and Aakanksha Chowdhery and Brian Ichter and Ayzaan Wahid and Jonathan Tompson and Quan Vuong and Tianhe Yu and Wenlong Huang and Yevgen Chebotar and Pierre Sermanet and Daniel Duckworth and Sergey Levine and Vincent Vanhoucke and Karol Hausman and Marc Toussaint and Klaus Greff and Andy Zeng and Igor Mordatch and Pete Florence},
  journal={arXiv preprint arXiv:2303.03378},
  year={2023}
}

@article{octo,
  title={Octo: An open-source generalist robot policy},
  author={Team, Octo Model and Ghosh, Dibya and Walke, Homer and Pertsch, Karl and Black, Kevin and Mees, Oier and Dasari, Sudeep and Hejna, Joey and Kreiman, Tobias and Xu, Charles and others},
  journal={arXiv preprint arXiv:2405.12213},
  year={2024}
}

@article{chai2025activation,
  title={Activation Reward Models for Few-Shot Model Alignment},
  author={Chai, Tianning and Mitra, Chancharik and Huang, Brandon and Gare, Gautam Rajendrakumar and Lin, Zhiqiu and Arbelle, Assaf and Karlinsky, Leonid and Feris, Rogerio and Darrell, Trevor and Ramanan, Deva and others},
  journal={arXiv preprint arXiv:2507.01368},
  year={2025}
}

@article{cunningham2023sae,
  title={Sparse Autoencoders Find Highly Interpretable Features in Language Models},
  author={Cunningham, Hoagy and Ewart, Aidan and Riggs, Logan and Huben, Robert and Sharkey, Lee},
  journal={arXiv preprint arXiv:2309.08600},
  year={2023},
  url={https://arxiv.org/abs/2309.08600}
}

@article{elhage2022superposition,
  title={Toy Models of Superposition},
  author={Elhage, Nicholas and Nanda, Neel and Olsson, Catherine and Geva, Mor and Reddy, Akbir Khan and others},
  journal={arXiv preprint arXiv:2209.10652},
  year={2022},
  url={https://arxiv.org/abs/2209.10652}
}

@article{gandelsman2023textspan,
  title={Interpreting CLIP's Image Representation via Text-Based Decomposition},
  author={Gandelsman, Yossi and Efros, Alexei A. and Steinhardt, Jacob},
  journal={arXiv preprint arXiv:2310.05916},
  year={2023},
  url={https://arxiv.org/abs/2310.05916}
}

@inproceedings{balasubramanian2024decompose,
  title={Decomposing and Interpreting Image Representations via Text in ViTs Beyond CLIP},
  author={Balasubramanian, Sriram and Basu, Samyadeep and Feizi, Soheil},
  booktitle={NeurIPS},
  year={2024}
}

@article{goh2021multimodalneurons,
  title={Multimodal Neurons in Artificial Neural Networks},
  author={Goh, Gabriel and Cammarata, Nick and Voss, Chelsea and Carter, Shan and Petrov, Michael and Schubert, Ludwig and Radford, Alec and Olah, Chris},
  journal={Distill},
  year={2021},
  doi={10.23915/distill.00030},
  url={https://distill.pub/2021/multimodal-neurons}
}

@article{basu2024mloc,
  title={On Mechanistic Knowledge Localization in Text-to-Image Generative Models},
  author={Basu, Samyadeep and Rezaei, Keivan and Kattakinda, Priyatham and Rossi, Ryan A. and Zhao, Nanxuan and Morariu, Vlad I. and Manjunatha, Varun and Feizi, Soheil},
  journal={arXiv preprint arXiv:2405.01008},
  year={2024},
  url={https://arxiv.org/abs/2405.01008}
}

@article{basu2024mlmstorage,
  title={Understanding Information Storage and Transfer in Multi-Modal Large Language Models},
  author={Basu, Samyadeep and Grayson, Martin and Morrison, Cecily and Nushi, Besmira and Feizi, Soheil and Massiceti, Daniela},
  journal={arXiv preprint arXiv:2406.04236},
  year={2024},
  url={https://arxiv.org/abs/2406.04236}
}

@book{Maslow1966-MASTPO,
	address = {New York,},
	author = {Abraham Harold Maslow},
	editor = {},
	publisher = {Harper \& Row},
	title = {The Psychology of Science},
	year = {1966}
}

@article{lin2025mmfm_survey,
  title={A Survey on Mechanistic Interpretability for Multi-Modal Foundation Models},
  author={Lin, Zihao and Basu, Samyadeep and Beigi, Mohammad and Manjunatha, Varun and Rossi, Ryan A. and Wang, Zichao and Zhou, Yufan and Balasubramanian, Sriram and Zarei, Arman and Rezaei, Keivan and others},
  journal={arXiv preprint arXiv:2502.17516},
  year={2025},
  url={https://arxiv.org/abs/2502.17516}
}

@article{schwettmann2023towards,
  title={Towards Mechanistic Interpretability of Multimodal Neurons},
  author={Schwettmann, Sarah and Watkins, Oliver and Wattenberg, Martin and Carter, Shan},
  journal={arXiv preprint arXiv:2301.11796},
  year={2023},
  url={https://arxiv.org/abs/2301.11796}
}

@inproceedings{
anwar2024contrast,
title={Contrast Sets for Evaluating Language-Guided Robot Policies},
author={Abrar Anwar and Rohan Gupta and Jesse Thomason},
booktitle={Conference on Robot Learning (CoRL)},
year={2024},
}

@article{taxonomy2025arxiv,
  title={A Taxonomy for Evaluating Generalist Robot Policies},
  author={Gao, Jensen and Belkhale, Suneel and Dasari, Sudeep and Balakrishna, Ashwin and Shah, Dhruv and Sadigh, Dorsa},
  journal={arXiv preprint arXiv:2503.01238},
  year={2025}
}

@article{liu2023tail,
  title={Tail: Task-specific adapters for imitation learning with large pretrained models},
  author={Liu, Zuxin and Zhang, Jesse and Asadi, Kavosh and Liu, Yao and Zhao, Ding and Sabach, Shoham and Fakoor, Rasool},
  journal={arXiv preprint arXiv:2310.05905},
  year={2023}
}

@inproceedings{liang2022transformer,
  title={Transformer adapters for robot learning},
  author={Liang, Anthony and Singh, Ishika and Pertsch, Karl and Thomason, Jesse},
  booktitle={CoRL 2022 Workshop on Pre-training Robot Learning},
  year={2022}
}

@article{xie2025data,
  title={Data Retrieval with Importance Weights for Few-Shot Imitation Learning},
  author={Xie, Amber and Chand, Rahul and Sadigh, Dorsa and Hejna, Joey},
  journal={Conference on Robot Learning (CoRL)},
  year={2025}
}

@article{lin2024flowretrieval,
  title={Flowretrieval: Flow-guided data retrieval for few-shot imitation learning},
  author={Lin, Li-Heng and Cui, Yuchen and Xie, Amber and Hua, Tianyu and Sadigh, Dorsa},
  journal={Conference on Robot Learning (CoRL)},
  year={2024}
}

@InProceedings{pmlr-v305-haon25a,
  title = 	 {Mechanistic Interpretability for Steering Vision-Language-Action Models},
  author =       {H\"{a}on, Bear and Stocking, Kaylene Caswell and Chuang, Ian and Tomlin, Claire},
  booktitle = 	 {Proceedings of The 9th Conference on Robot Learning},
  pages = 	 {2743--2762},
  year = 	 {2025},
  editor = 	 {Lim, Joseph and Song, Shuran and Park, Hae-Won},
  volume = 	 {305},
  series = 	 {Proceedings of Machine Learning Research},
  month = 	 {27--30 Sep},
  publisher =    {PMLR},
  url = 	 {https://proceedings.mlr.press/v305/haon25a.html}
}

@software{jax2018github,
  author = {James Bradbury and Roy Frostig and Peter Hawkins and Matthew James Johnson and Chris Leary and Dougal Maclaurin and George Necula and Adam Paszke and Jake Vander{P}las and Skye Wanderman-{M}ilne and Qiao Zhang},
  title = {{JAX}: composable transformations of {P}ython+{N}um{P}y programs},
  url = {http://github.com/jax-ml/jax},
  version = {0.3.13},
  year = {2018},
}

@article{beyer2024paligemma,
  title={Paligemma: A versatile 3b vlm for transfer},
  author={Beyer, Lucas and Steiner, Andreas and Pinto, Andr{\'e} Susano and Kolesnikov, Alexander and Wang, Xiao and Salz, Daniel and Neumann, Maxim and Alabdulmohsin, Ibrahim and Tschannen, Michael and Bugliarello, Emanuele and others},
  journal={arXiv preprint arXiv:2407.07726},
  year={2024}
}

@article{steiner2024paligemma,
  title={Paligemma 2: A family of versatile vlms for transfer},
  author={Steiner, Andreas and Pinto, Andr{\'e} Susano and Tschannen, Michael and Keysers, Daniel and Wang, Xiao and Bitton, Yonatan and Gritsenko, Alexey and Minderer, Matthias and Sherbondy, Anthony and Long, Shangbang and others},
  journal={arXiv preprint arXiv:2412.03555},
  year={2024}
}

@misc{Polymetis2021,
  author =       {Lin, Yixin and Wang, Austin S. and Sutanto, Giovanni and Rai, Akshara and Meier, Franziska},
  title =        {Polymetis},
  howpublished = {\url{https://facebookresearch.github.io/fairo/polymetis/}},
  year =         {2021}
}
